\pdfoutput=1
\def\year{2016}\relax
\documentclass[letterpaper]{article}
\usepackage{lmodern}
\usepackage{styl}
\usepackage[linktoc=all]{hyperref}
\usepackage{times}
\usepackage{helvet}
\usepackage{courier}
\let\chapter\section
\usepackage[ruled,vlined]{algorithm2e}
\usepackage{amsthm}
\usepackage{graphicx}
\usepackage{booktabs}
\usepackage [english]{babel}

\addto\captionsenglish{%
}
\usepackage [autostyle, english = american]{csquotes}
\MakeOuterQuote{"}
\usepackage[skip=10pt]{caption}
\usepackage{natbib}
\begin{document}

\title{Fitted Learning: Models with Awareness of their Limits}

\author{Navid Kardan \textnormal{and} Kenneth O. Stanley\\
Department of Computer Science, University of Central Florida,\\ 
4328 Scorpius Street,\\ 
Orlando, FL 32816-2362,\\
kardan@knights.ucf.edu, kstanley@cs.ucf.edu
}
\maketitle
\begin{abstract}
	Though deep learning has pushed the boundaries of classification forward, in recent years hints of the limits of standard classification have begun to emerge. Problems such as fooling, adding new classes over time, and the need to retrain learning models only for small changes to the original problem all point to a potential shortcoming in the classic classification regime, where a comprehensive a priori knowledge of the possible classes or concepts is critical.  Without such knowledge, classifiers misjudge the limits of their knowledge and overgeneralization therefore becomes a serious obstacle to consistent performance. In response to these challenges, this paper extends the classic regime by reframing classification instead with the assumption that concepts present in the training set are only a sample of the hypothetical final set of concepts.  To bring learning models into this new paradigm, a novel elaboration of standard architectures called the \emph{competitive overcomplete output layer} (COOL) neural network is introduced. Experiments demonstrate the effectiveness of COOL by applying it to fooling, separable concept learning, one-class neural networks, and standard classification benchmarks. The results suggest that, unlike conventional classifiers, the amount of generalization in COOL networks can be tuned to match the problem. 

\end{abstract}
\section{Introduction}

The hope in machine learning is often to match or even exceed the performance of humans in tasks once prohibitive to machines.  The measure of such success is frequently expressed as the \emph{accuracy} of the learning algorithm on a test set separate from the examples experienced during training.  For example, in ILSVRC-2010 (a subset of ImageNet), a dataset with roughly $1.2$ million images spanning one thousand of different classes, success is measured on a separate $150$,$000$ images in the test set \citep{imagenet}.  This conventional approach to measuring success then makes it possible to compare machine and human performance if we assess humans on the same test set.  In fact, in ImageNet humans achieve about $94.9\%$ on the test set \citep{imagenet}, a level recently exceeded by deep learning algorithms \citep{resnet}.  Thus a narrative emerges whereby machines are creeping past the abilities of humans in visual recognition, a remarkable proposition.

Yet this potential narrative of course oversimplifies the depth of human comprehension hidden beneath the raw accuracy score.  Unlike machines, humans \emph{have a sense of the limits of their own knowledge}, a fact not easily exposed through a fixed test set drawn from a similar distribution as the training examples.  For example, confronted with adversarially-selected images reminiscent of random noise, a human can easily recognize that the resultant image is not familiar and hence belongs to no known class at all.  Deep learning, in contrast, has proven easily fooled by such images into assigning them to a class with over $99\%$ confidence, a phenomenon called fooling \citep{szegedy,nguyen}.   

While being fooled by examples far outside the training distribution raises concerns, such mistakes alone are confined to scenarios that might be dismissed as relatively unlikely.  For example, how often in the real world might one encounter an image unlike anything in the real world yet adversarially optimized to trigger a mistaken classification?  However, as experiments in this paper will show, fooling is only a foreshadowing of a set of shortcomings in conventional neural networks that fall short of human-level capabilities.  For example, a tendency to classify images with confidence that are not from among the training classes means that adding a new class to an already-learned model is unlikely to work, thereby precluding continual learning of new classes.  It also means that combining separately-trained classifiers is similarly untenable.  Single-class neural networks, which learn to say whether a training instance is within the single class or outside it, are also prohibitive to train for similar reasons. 

All these problems result from the tendency of conventional neural networks to draw well-informed discriminative borders between different classes, but to fail at confining the regions of such classes to the span only of the examples in the dataset.  The consequences include fooling, but also real practical divergence from the natural ability to learn new classes seamlessly throughout a lifetime.  If neural networks could be brought to learn better fitted regions, i.e.\ ones that do not \emph{overgeneralize}\footnote{In psychology, in early language acquisition, the same concept is called \emph{overextension} and is often attributed to a limitation in vocabulary \citep{overextension}.}, the problems so far cited, from fooling to continual learning, would be largely mitigated or even in some cases solved.

To capture this idea this paper later introduces a formal framework for \emph{fitted learning}, which refers to algorithms that learn both good discriminative borders and the distribution of training data at the same time.  Moreover, the major contribution of this work is a simple mechanism for achieving fitted learning (and to tune the level of fit) within the conventional apparatus of deep learning and neural networks.  The main idea, called the \emph{competitive overcomplete output layer} (COOL), is to assign more than one output to each class and also force such outputs to compete with each other to respond correctly.  The resulting dynamic, which is achieved through the usual process of stochastic gradient descent, pushes the competing neurons within a class to respond to overlapping yet non-identical regions of input space, yielding sufficient consensus for confident classification only within a tight-fitting region around the training data.   This simple mechanism for fitted learning is thereby a step towards a more natural style of learning in neural networks.

Experiments in this paper in the MNIST digit-classification domain \citep{mnist} confirm that the COOL approach not only mitigates fooling as expected, but also indeed enables combining multiple separately-learned classifiers into one, and even makes possible effective one-class neural networks, all of which are otherwise prohibitive with conventional convolutional neural networks (CNNs) \citep{mnist}.  

After covering background material next, the paper introduces the components of the COOL architecture and then intuitively demonstrates their advantage through visualizations of COOL and non-COOL networks in a simple two-dimensional classification problem.  The COOL mechanism is then justified more formally and tested within the context of fooling (false positive) images , after which fitted learning and the connected \emph{generalized classification problem} are also formalized.  A series of experiments in MNIST follow on learning separate classifiers that are later combined (which works poorly with conventional CNNs), including the first demonstration of working one-class neural networks.  The experiments then conclude with head-to-head classification comparisons in CIFAR-10 and CIFAR-100, followed by a discussion of the implications of fitted learning and COOL. 
\section{Related Work}
A close conceptual framework to fitted learning is the \emph{open set recognition problem} (OSRP) \citep{openset,Jain2014} in the vision community, where unknown classes may appear in the testing phase and the learning model should be able to detect such unknowns. However, OSRP focuses on related classes to the domain of the problem, while 
the aim in fitted learning is for learning models to identify all possible unfamiliar examples.
That is, a fitted model should be able to discriminate at the same time as capturing the data distribution for each class. Unlike the common practice in open set recognition, a fitted model should learn a tight halo around different class regions, unless there is a border friction with another class, and therefore should not be very sensitive to the choice of a threshold parameter. Visualization experiments that follow indeed illustrate these properties for COOL in low-dimensional spaces.

An established method that is applied in OSRP and anomaly detection (amongst others) is the one-class SVM \citep{oneclasssvm}, which tries to capture the data distribution, or more technically the support of the probability distribution, by turning the one-class problem into a classification problem. The main difference between the one-class SVM and fitted models is that the one-class SVM is a discriminative model at its core so it will find a border around some data points rather than capturing their underlying distribution. For example, learning regions in the feature space with a hollow inside is very difficult (if not impossible) for a one-class SVM model. 

A different way to address unfamiliar examples is by outputting an uncertainty value for every prediction that the model makes. Recently \citet{gal} proposed a procedure to extract such uncertainty values in deep models using Bayesian neural networks.  

Fitted learning can also be viewed as inverse generative model learning, where sampling from a particular class involves searching for an input instance that highly activates the corresponding class in the hypothetical fitted model. That is, an input that highly activates an output node in a trained fitted model is an example of the class of that output node and a procedure to generate such instances leads to a generative model. An analogy can be drawn with Generative Adversarial Nets (GANs) \citep{gan}, which try to capture the distribution of data through a generative model that competes against a discriminative model. COOL, as a potential fitted model, induces some form of competition as well, but unlike GANs it does not entail careful training of the competing models because there are no explicit separate models. 

Architecture-wise, COOL applies overcomplete output layers. The well-known Inception \citep{inception} architecture also applies two softmax output layers to train a deep NN, which could be interpreted as another example of overcomplete output layers.  One of these output layers is attached to a hidden layer with a modest depth that (together with the main output layer) would provide the network with appropriate gradient signals for training the network in the early stages of learning. This extra output layer is later detached from the network and the main output layer located in a higher depth takes over and continue training. 

In this setup, the Inception network is essentially assigning two output units to each class. However, COOL does not introduce two output layers. Moreover, unlike COOL, the second output layer in Inception is only applied temporarily to alleviate the gradient vanishing effect and has no effect in the actual decision making of the network in the test phase.

Another architecture form literature with similarities to COOL is the multiverse loss network \citep{multiverse} that contains several softmax output layers with an orthogonality constraint between them. The authors applied this network to get a more diverse set of \emph{features} that are then successfully applied for transfer learning \citep{multiverse}. Again this network never utilizes its output layers, but rather just later leverages its final hidden units to obtain diverse features for the transfer learning task.

Another method with a conceptual connection to COOL is dropout \citep{dropout2}, where an exponential number of models are trained together by randomly dropping out some hidden units during the training phase. Similarly, COOL in effect trains an exponential number of models as well, which also try to learn the same concept. However, COOL also employs a mechanism to encourage diversity in the final subset models and in contrast to dropout it usually results in the speedup of training.


\section{Approach}
COOL in effect expands the traditional output layer by assigning several output units to each class or concept. Furthermore, forcing this group of neurons to compete ensures their diversity by partitioning the input space among output units of the same class. As a result these output units ultimately only agree within regions proximal to the training instances. In other words, unlike conventional architectures, COOL captures the distribution of the training instances of each class \emph{at the same time} as learning to discriminate between the instances of different classes. 

The section begins with a description of the basic unit of COOL, the \emph{neuron aggregate}. The remainder of the section then describes the proposed new architecture for output layers.
\subsection{Neuron Aggregate}
A \emph{neuron aggregate} is a collection of units, called \emph{member units}, that is intended to learn a single concept. In this sense, a neuron aggregate acts as an ensemble and is inspired by the hypothesis (in contrast to the "grandmother cell") that in biological neural networks more than one neuron is involved in recognizing each concept \citep{grandma,neuroscience}.

An \emph{internally-competitive aggregate} is a neuron aggregate whose member units' activations are mutually inhibitory (e.g.~through reciprocal inhibition) and is modeled in COOL by a softmax function applied to member units. Note that it accordingly suffices for a neuron aggregate to be part of a softmax layer to become an internally-competitive aggregate. 

\subsection{The Competitive Overcomplete Output Layer (COOL)}

A COOL consists of a simple change to the traditional output layer setup of neural networks. In particular, each output unit is replaced by an internally-competitive aggregate. As we elaborate further on this architecture, in the rest of this paper we assume one-hot neural encoding for target output labels. We also define the \emph{expected maximum activation value (EMAV)} of a unit as the expected\footnote{The word "expected" conveys that maximum activation value a unit can take during training may differ from the maximum value it can output during testing.} highest activation value a single unit can have; for example, normally a traditional logistic sigmoid saturates at an EMAV of $one$. An output unit in a softmax layer normally is trained with values of either $0$ or $1$, indicating the deactivation/activation of the corresponding unit, respectively. In our framework, because a traditional unit is replaced by a neuron aggregate, COOL accepts lower values to indicate the activation of member units. More specifically, $1/\omega$ indicates activation of a member unit and zero indicates its deactivation, where $\omega$ is a hyperparameter called the \emph{degree of overcompleteness (DOO)} that indicates the number of member units in a neuron aggregate. In this paper all neuron aggregates of the same output layer use the same degree of overcompleteness. In effect, the EMAV value of each member unit of an aggregate is $1/\omega$.

During the training phase, all the member units of the same neuron aggregate are trained with the same value, i.e.~zero or $1/\omega$, depending on the desired activation of the corresponding aggregate. Algorithm \ref{alg1} shows this procedure formally. Note that $1/\omega$ combined with the one-hot encoding of class labels ensure that the sum of all the member units always equals $1.0$; a cross-entropy cost function is then applied straightforwardly to train such a network. The activation values of member units in a COOL setup can be interpreted as the scaled probability of inclusion of the input instance in a particular class.

It is important to highlight the role of the two components of a COOL layer, i.e.~overcompleteness and competitive training through the softmax function. The first component in effect adds new dimensions to the output layer while the second enforces a continuing competition among the member units of the active neuron aggregate. The section on \emph{Why the COOL Mechanism Works} shows in detail how these two components enforce a beneficial competition among member units. Visualization experiments that come first provide initial intuition on how this competition can lead to a fundamental change in output layer behavior, in effect preventing output units from overgeneralizing into regions far away from the training instances. These experiments will suggest that internally-competitive aggregates are able to learn an appropriate probability distribution of their assigned concepts, a critical property for many learning tasks.

During the inference phase, the outputs of the member units of an aggregate are combined through \emph{multiplication} to calculate the total activation value of the aggregate. The idea is that member units are trained to minimize variance and therefore are expected to exhibit the least variance and highest activation when the network inputs are within the proximity of training instances. At the same time, softmax enforces a constant total sum of the member units. Figure~\ref{cool} compares a conventional output layer with a COOL, including the multiplications that happen for inference.

\begin{figure}[!t]
	\begin{minipage}{0.32\columnwidth}
		\centering
		\centerline{\includegraphics[width=0.95\columnwidth]{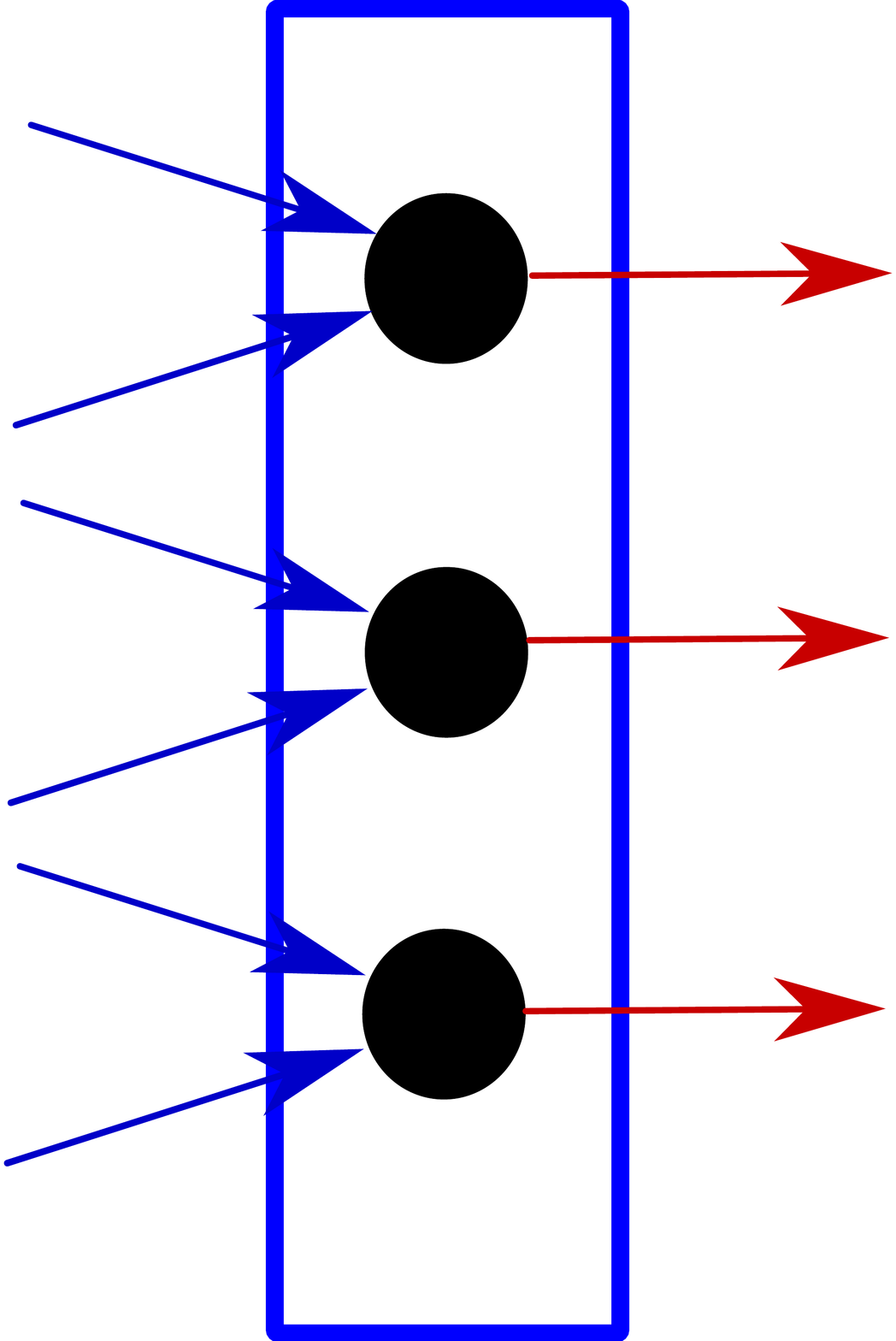}}
		\captionsetup{labelformat=empty}
		\caption{(a) Conventional\\ output layer}
	\end{minipage}%
	\begin{minipage}{0.32\columnwidth}
		\centering
		\centerline{\includegraphics[width=0.95\columnwidth]{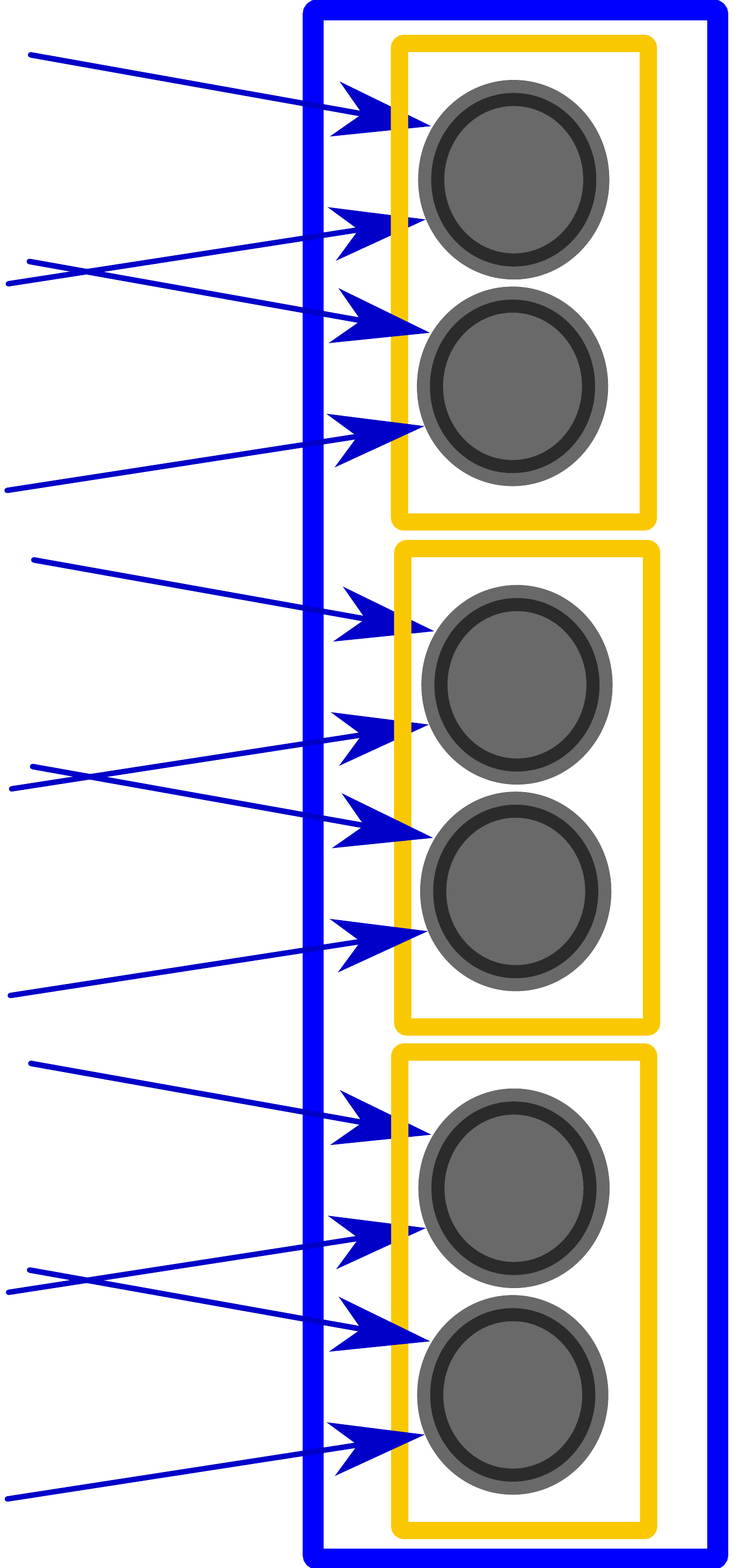}}
		\captionsetup{labelformat=empty}
		\caption{(b) COOL \\during training}
	\end{minipage}%
	\begin{minipage}{0.32\columnwidth}
		\centering
		\centerline{\includegraphics[width=0.95\columnwidth]{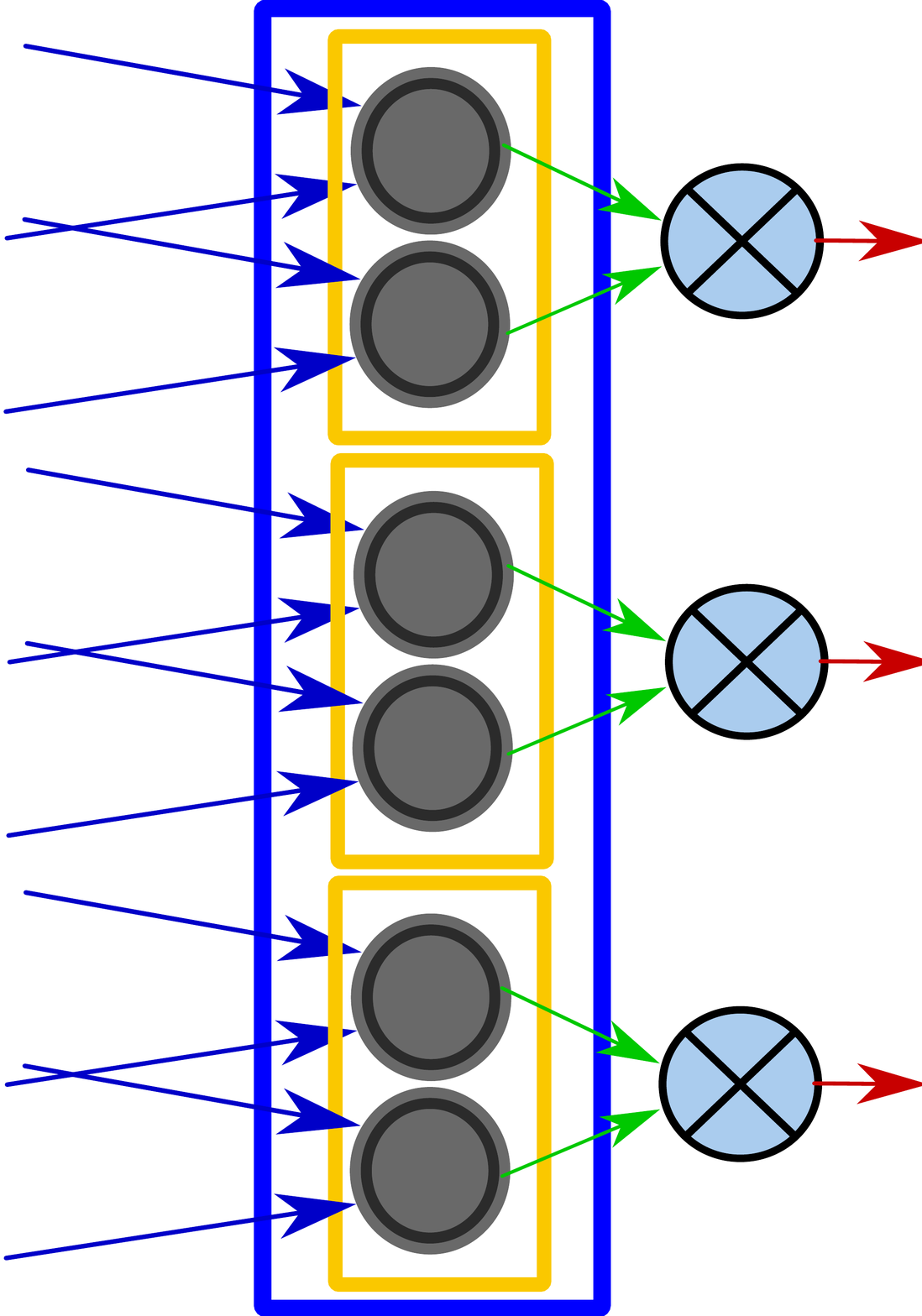}}
		\captionsetup{labelformat=empty}
		\caption{(c) COOL \\during testing}
	\end{minipage}
	\setcounter{figure}{0}
	\caption{\textbf{The Output layer in a conventional neural network vs.\ COOL.} (a) A conventional output layer has three neurons to learn three concepts. (b) An overcomplete output layer for the same task is shown during training. Each yellow rectangle depicts an aggregate and two gray units in each aggregate show that all aggregates have DOO of $2$ in this architecture. (c) The same COOL output layer computes its decisions through multiplication during the test phase.}
	\label{cool}
\end{figure}

Multiplication helps to maximize activation (and minimize variance) of member units at low computational cost because the product of a set of variables constrained by a finite sum is maximized when the variables are all equal. Finally, to control the degree of generalization of a particular class $c$, the output values can be exponentiated by a softness parameter $\sigma_c$ after application of the multiplication. This operation allows tuning the decision boundaries of individual classes arbitrarily. The role of this parameter will be more clear when separable concept learning is later revealed as a consequence of COOL. Optionally, to get final results in the form of probability values each aggregate output can be multiplied by $\omega^\omega$. Algorithm \ref{alg2} describes the procedure for calculating the activation value of an aggregate during testing. 

It is worth noting that if multiplication is replaced by addition, COOL will in effect work similarly to a conventional neural network. However, having several output units per class makes even COOL with addition advantageous in some situations.
For example, the training may still converge with fewer iterations, or we might be able to train bigger architectures with less overfitting.

\begin{algorithm}[!t]
	\KwIn{degree of overcompleteness $\omega$,
		class index for a training instance $\eta$,
		total number of classes $k$}
	\KwOut{training target $l$}
	$l$ $\gets$ zeros($\omega\times k$) 
	
	\For{$i \gets 1$ \textbf{to} $\omega$}{
		$l[(i-1)\times k + \eta +1] \gets \frac{1}{\omega}$
	}

	\caption{How the target for a training instance is set in COOL. }
	\label{alg1}
\end{algorithm}

\begin{algorithm}[!t]
	\KwIn{degree of overcompleteness $\omega$,
		prediction vector $p$,
		softness parameter vector $\sigma$,
		number of classes $k$}
	\KwOut{test label probability vector $l$}
	$l \gets ones(k)$
	
	\For{$i \gets 1$ \textbf{to} $\omega$}{
		\For{$j \gets 1$ \textbf{to} $k$}{
			$l[j] \gets l[j] \times p[j + (i-1)\times k] \times \omega$ 
		}
	}
	\For{$j \gets 1$ \textbf{to} $k$}{
		$l[j] \gets l[j]^{\sigma_j}$ 
	}

	\caption{Inferring a test instance in COOL. }
	\label{alg2}
\end{algorithm}

\section{Visualization Experiments}
\label{vis}
In this section a low-dimensional artificial dataset helps to visualize the behavior of COOL networks and compare them with traditional neural networks. These experiments suggest that COOL networks are able to successfully capture the underlying distribution of training instances while still preserving the generalization ability of conventional neural networks. In other words, \emph{they construct a notion of the limits of their knowledge}; they can recognize instances that are not inside their scope of expertise (i.e.~far from the training points) in addition to behaving like a traditional learning model. 
\subsection{Problem Definition}
\label{twocircle}
A major obstacle to understanding neural networks is the inability to visualize their behavior over a large cross-section of the input space. To address this challenge, a simple two-dimensional classification problem is introduced that allows such broad visualization. The problem involves classifying points that are located only within the perimeters of two concentric circles of points in the instance space, $c_1$ and $c_2$, with different radii and both centered at $(0, 0)$. The training set includes only points within these two bands (dashed circles in figure~\ref{trset}). Each point within circle $c_1$ (inner circle) is a member of class one and each point within circle $c_2$ (outer circle) is a member of class two. A uniform sampling of these circles generates the training set.

\begin{figure}[!b]
	\vskip 0.1in
	\begin{center}
		\centerline{\includegraphics[width=\columnwidth]{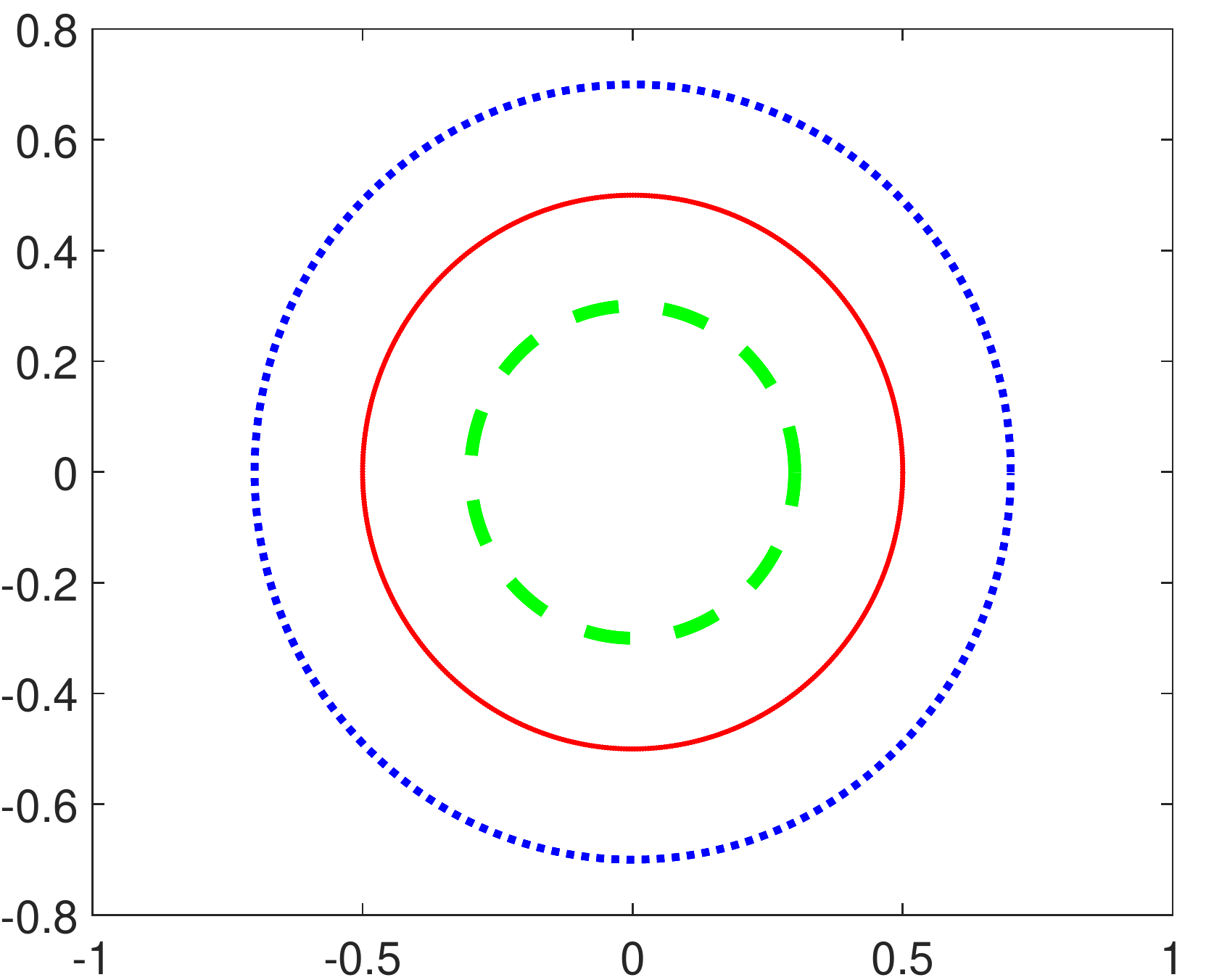}}
		\caption{\textbf{Illustration of the training data for the visualization problem.} The blue and green dots show the instances from two different classes and the red circle depicts the max-margin border.}
		\label{trset}
	\end{center}
	\vskip -0.1in
\end{figure}
Visualization experiments in this section train neural networks with different architectures on this two-class problem. The trained network's output for all possible two-dimensional instances at a high resolution is then plotted on a grid. That way, it is possible to see how the network classifies instances outside the training distribution.

Traditional neural networks and other discriminative classification models look for a discriminative border. For a two-class problem this objective means that they will typically partition the whole input space into two regions, both assigned to one of the available classes. One possible choice for such partitioning is a max-margin border, which is depicted in figure~\ref{trset} as a red line.

In the first experiment a shallow architecture (one hidden layer of 200 neurons) is compared with a deep one, consisting of three hidden layers of 400, 300, and 200 neurons, respectively (neither with COOL). Figure~\ref{shallowvsdeep} depicts the activation map of the output unit assigned to the inner circle in both the shallow and deep neural networks. These pictures show how adding more hidden layers result in finding a better decision boundary. However, the deep approach also tends to preserve high activation of output units in the regions that are far from the training points, a problem that can be described as \emph{overgeneralization}. In classification problems where every possible input instance is a member of the set of known classes overgeneralization does not exist by definition, but in most real-world applications:
\begin{itemize}
	\item the complete set of classes is \emph{not} known and 
	\item not all the possible inputs are an instance of \emph{any} class.   
\end{itemize}  

\begin{figure*}[!t]
	\centering
	
	\setcounter{figure}{0}	
	\begin{minipage}{0.5\textwidth}
		\centering
		\centerline{\includegraphics[width=\columnwidth]{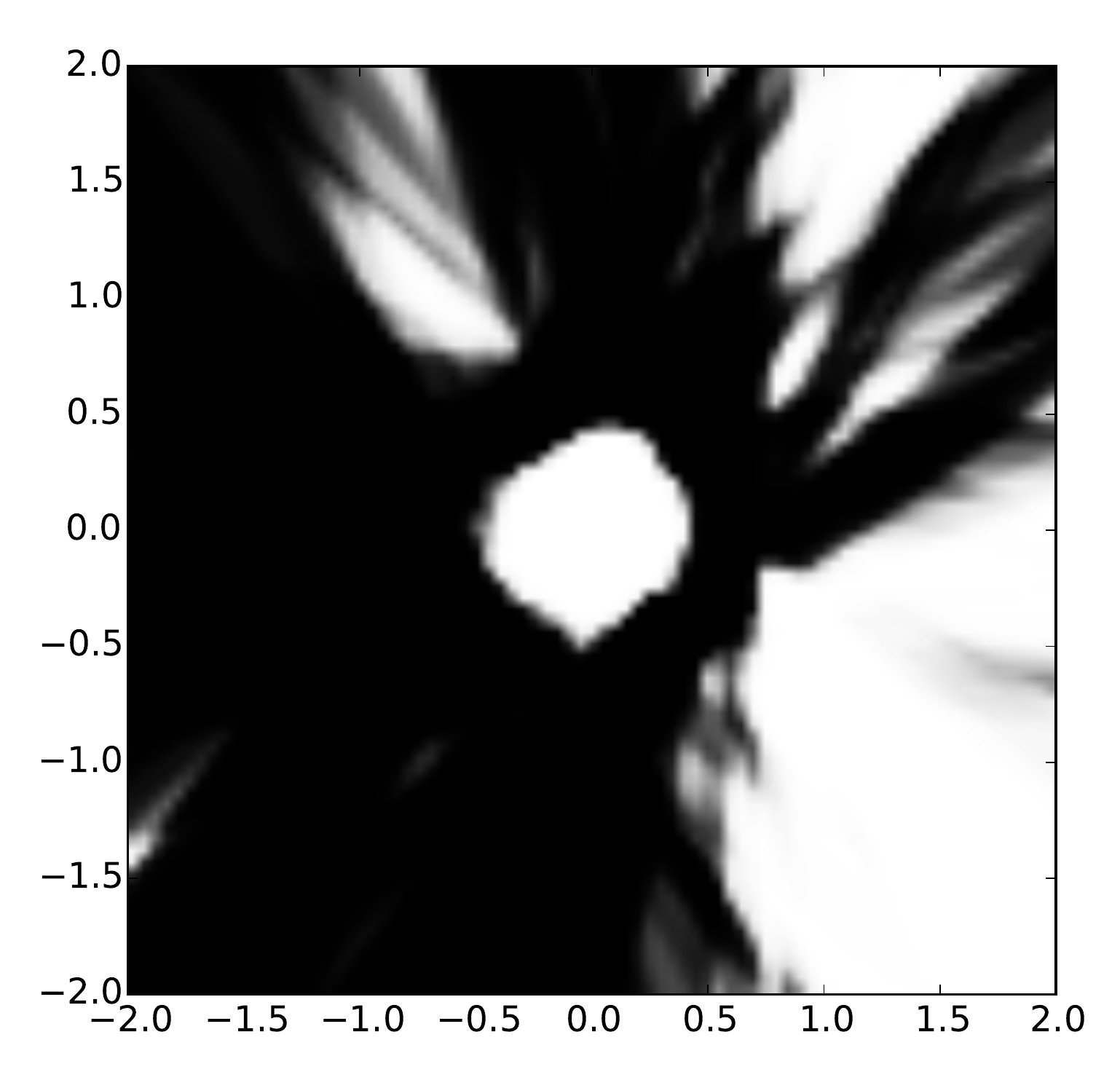}}
		\captionsetup{labelformat=empty,skip=0pt}
		\caption{(a) Shallow network}
	\end{minipage}%
	\begin{minipage}{0.5\textwidth}
		\centering
		\centerline{\includegraphics[width=\columnwidth]{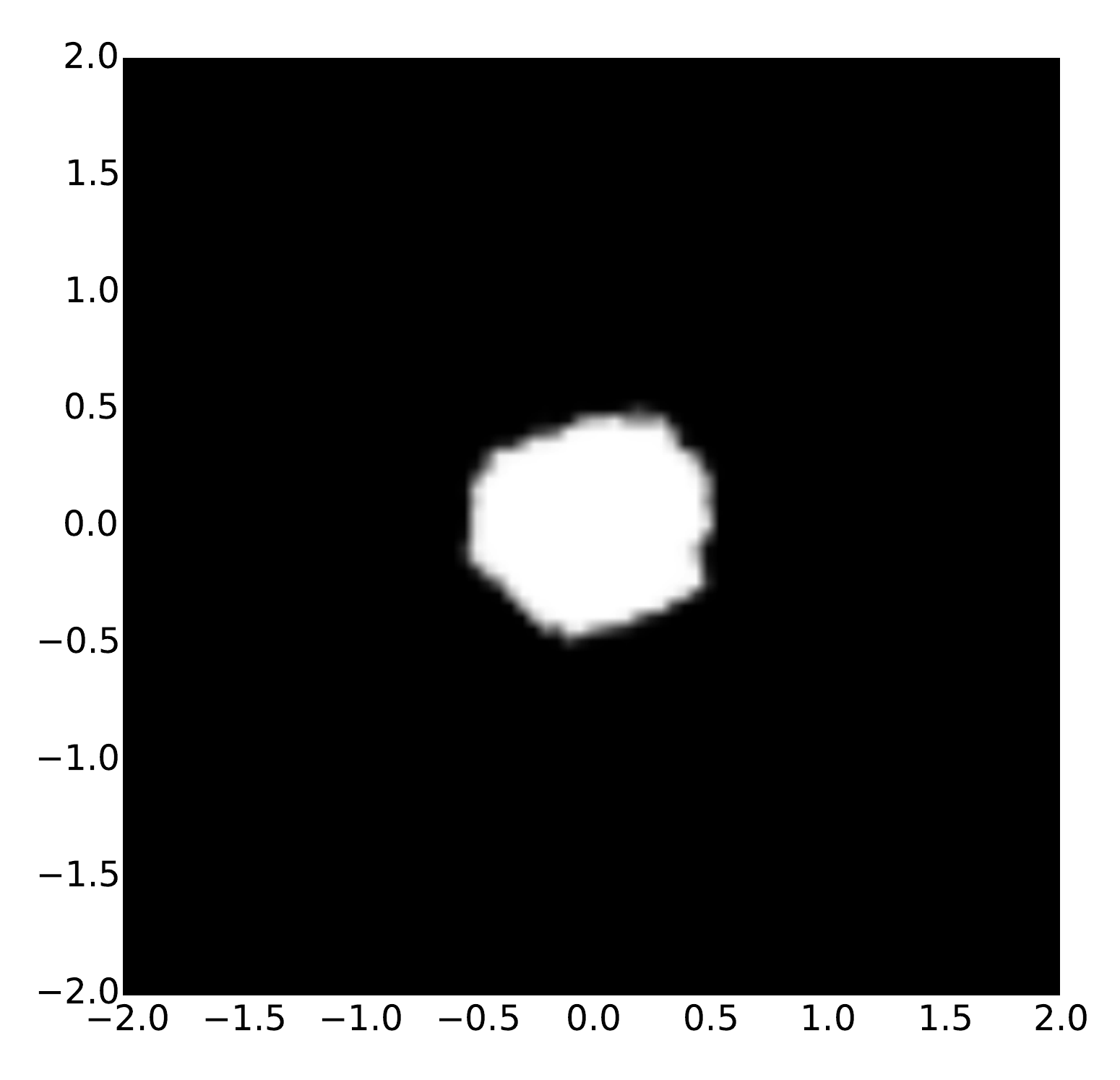}}
		\captionsetup{labelformat=empty,skip=0pt}
		
		\caption{(b) Deep network}
	\end{minipage}
	\caption{\textbf{Generalization ability of conventional deep vs.\ shallow neural networks on the visualization problem.} The activation map of the output unit assigned to learn the points within the inner circle is shown for both networks. (The activation map of the outer circle, which is not shown, is the complement of these plots.) The highly-activated area inside the circle in both plots indicates the ineffectiveness of conventional neural networks in preventing overgeneralization. (a) The abundance of highly activated points outside the inner circle is usually interpreted as a shortcoming of shallow neural networks in generalizing well. (b) The activation map of the deep architecture finds a better approximation of the max-margin decision border. While this outcome is conventionally considered a good behavior, it also hints that deep models are easier to be fooled.}
	\label{shallowvsdeep}
\end{figure*}

In fact, \emph{most} possible inputs may not be valid and only those input instances that are likely to be drawn from the training data distribution should be considered valid. One naive solution to the overgeneralization problem is to add an extra class including those inputs that are not a member of any class. There are two major drawbacks with this approach:
\begin{itemize}
	\item Accurate generation of such instances necessitates access to the underlying training instance generation process.
	\item Usually this set of instances has a much larger size, which can result in a dramatic data imbalance.
\end{itemize} 
In contrast, COOL provides an alternative solution that does not require any modification to the training data. To demonstrate this capability, in the next visualization a COOL network is trained on the same two-circle problem. Figure~\ref{rings} depicts the activation maps of the two output units corresponding to inner and outer circles, respectively (for both a COOL network and conventional network for comparison). These pictures clearly show how the COOL mechanism can effectively prevent overgenaralization, suggesting the ability of these types of network to capture an \emph{implicit} understanding of the data generation process. 

\begin{figure}[!t]
	\begin{minipage}{0.497\columnwidth}
		\centering
		\centerline{\includegraphics[width=0.95\columnwidth]{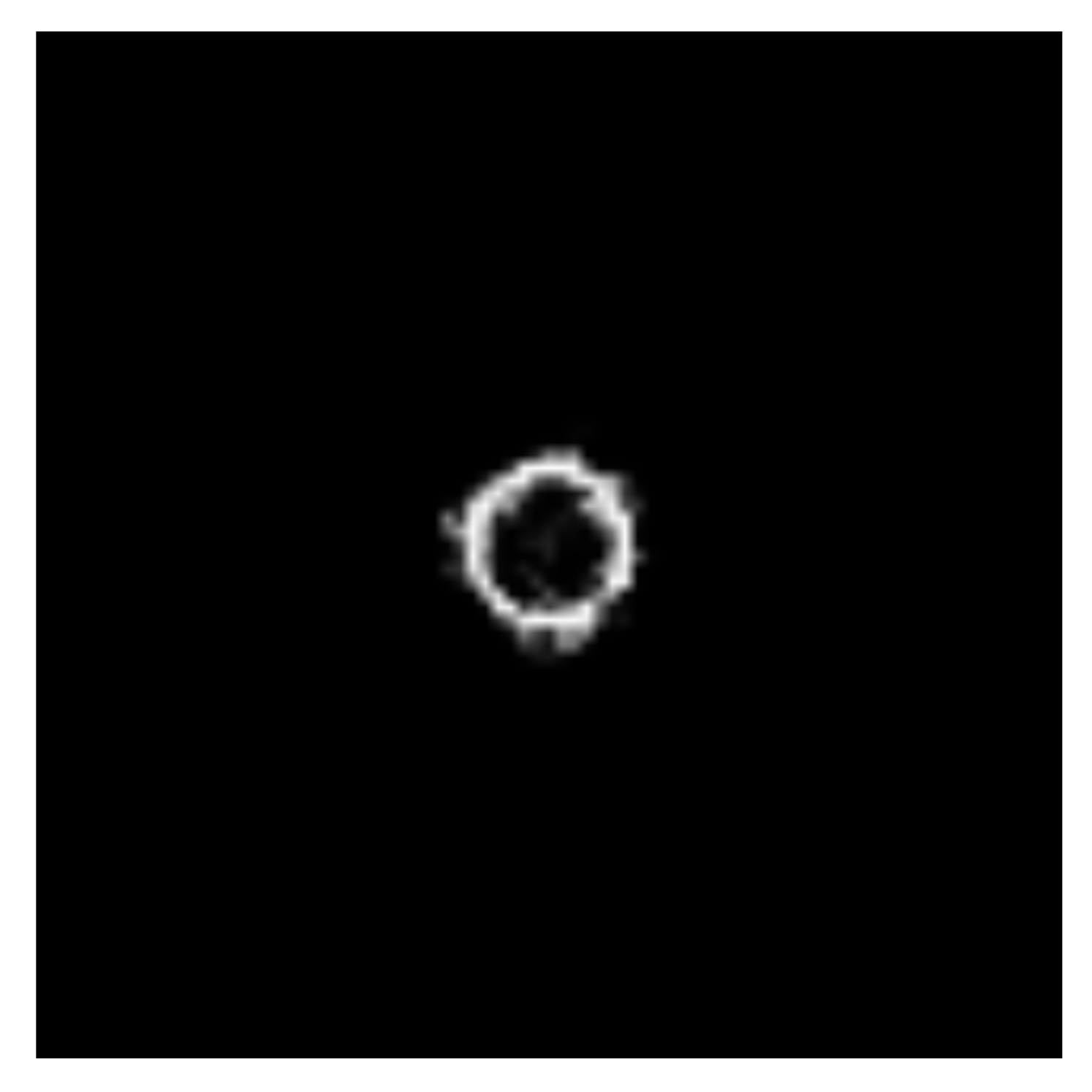}}
		\captionsetup{labelformat=empty,skip=7pt,labelsep=newline,singlelinecheck=false}
		\caption{(a)~COOL inner circle \\output}
	\end{minipage}%
	\begin{minipage}{0.50\columnwidth}
		\centering
		\centerline{\includegraphics[width=0.95\columnwidth]{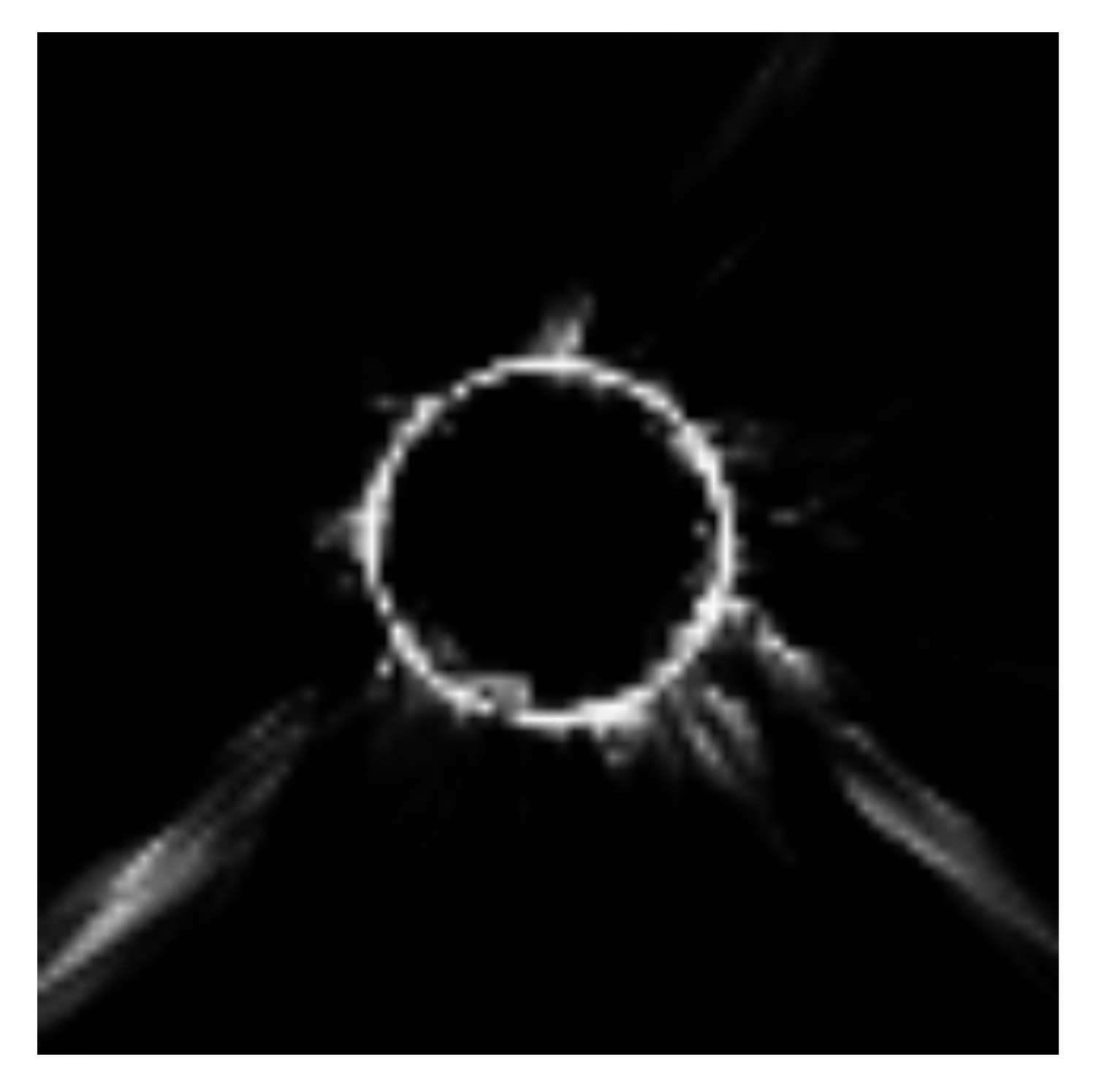}}
		\captionsetup{labelformat=empty,skip=7pt,labelsep=newline,singlelinecheck=false}
		\caption{(b)~COOL outer circle \\output}
	\end{minipage}
	\setcounter{figure}{1}		
	\begin{minipage}{0.5\columnwidth}
		\centering
		\centerline{\includegraphics[width=0.95\columnwidth]{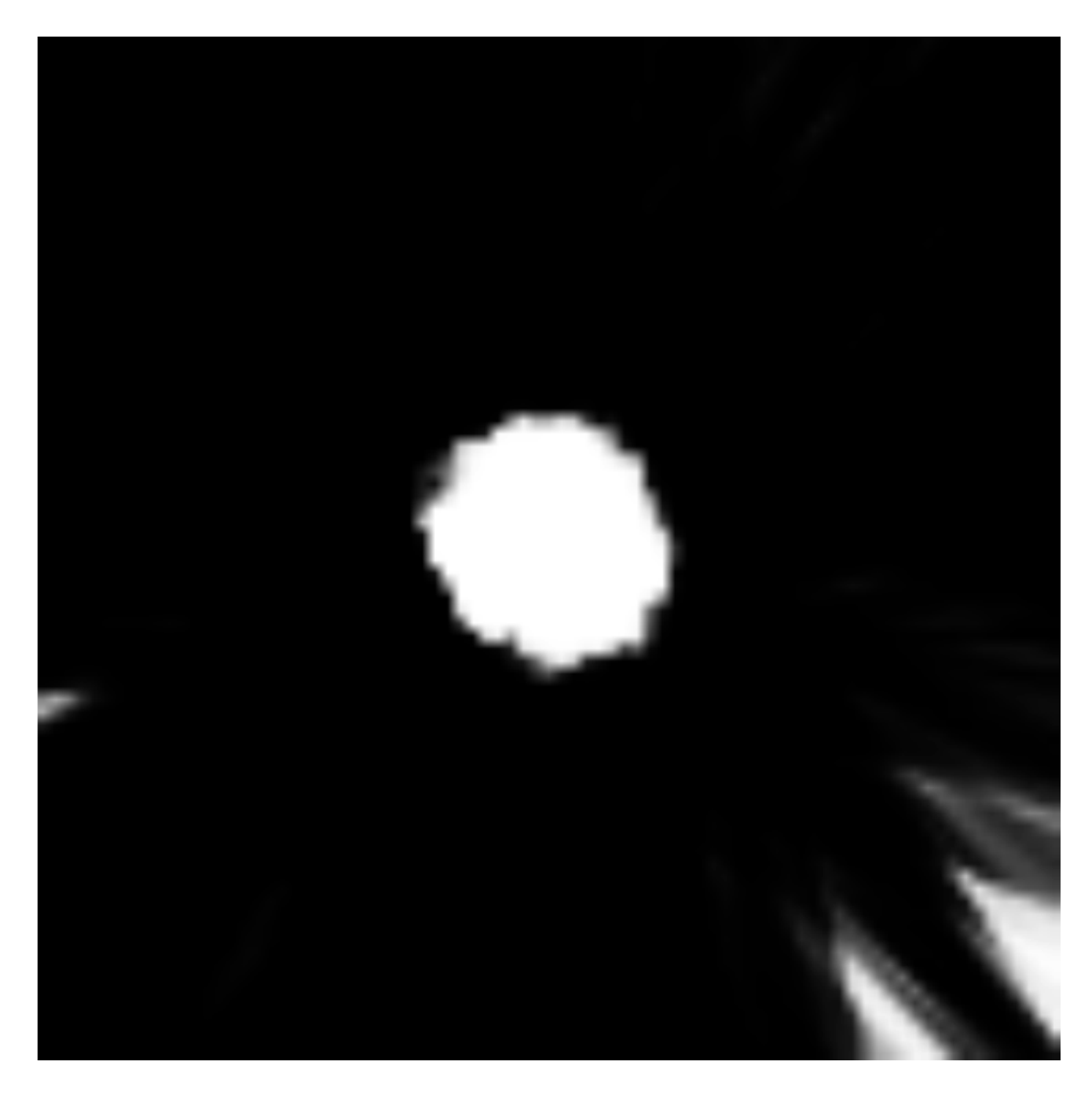}}
		\captionsetup{labelformat=empty,skip=7pt,labelsep=newline,singlelinecheck=false}
		\caption{(c)~Conventional neural \\network inner circle output}
	\end{minipage}%
	\begin{minipage}{0.5\columnwidth}
		\centering
		\centerline{\includegraphics[width=0.95\columnwidth]{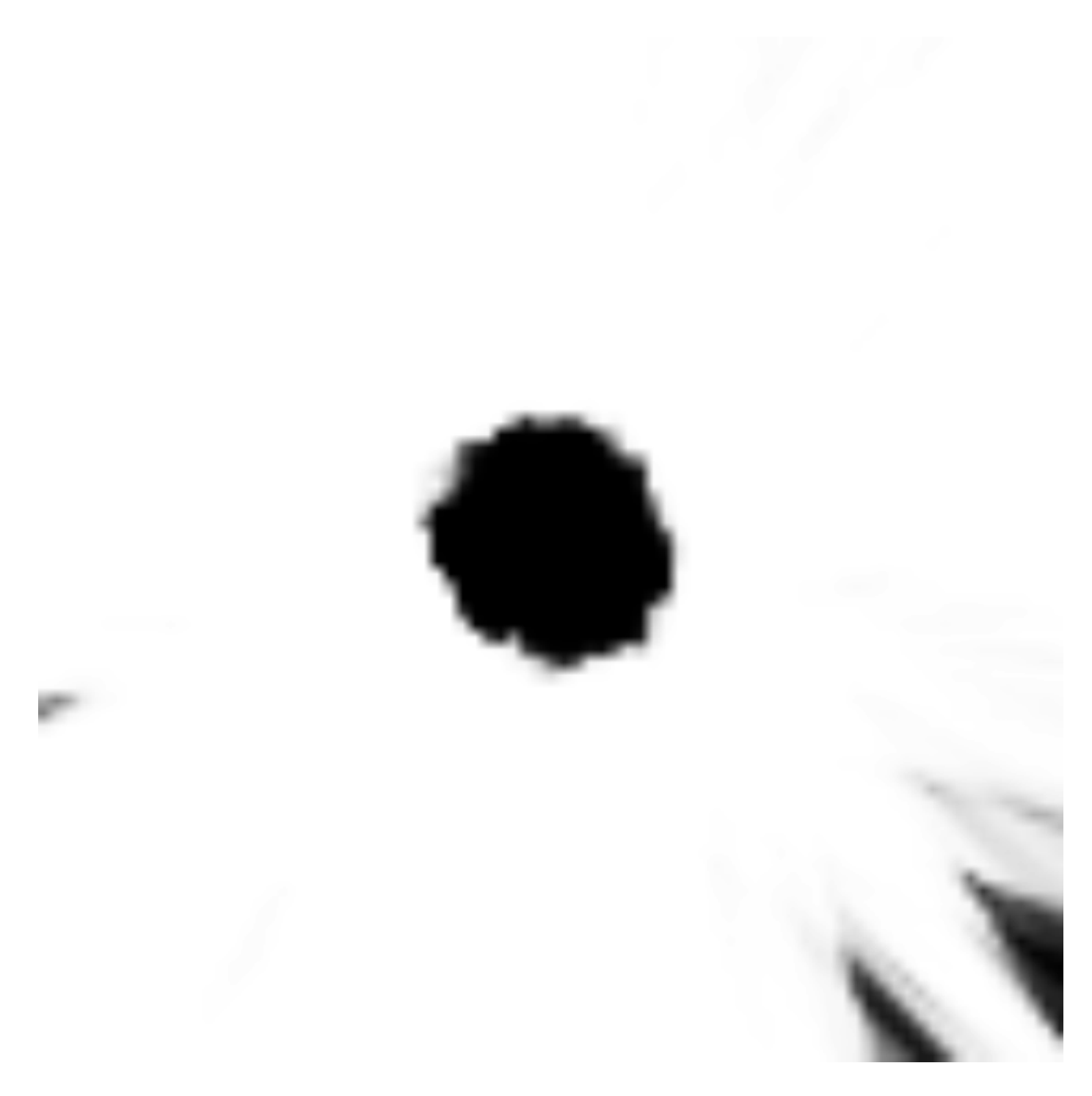}}
		\captionsetup{labelformat=empty,skip=7pt,labelsep=newline,singlelinecheck=false}
		\caption{(d)~Conventional neural \\network outer circle output}
	\end{minipage}
	\caption{\textbf{The activation map of output units in a COOL vs.\ a conventional neural network (hollow circle is ideal).} Images (a) and (c) depict the highly active area for the inner circle in a COOL versus a conventional MLP. Notice the presence of the donut hole in (a). Images (b) and (d) show the highly active region for the outer circle. Again the highest outputs for COOL are within the correct band around training data while the conventional neural network generalizes to infinity. These plots depict how COOL can prevent overgeneralization at the same time as learning to solve the problem.}
	\label{rings}
\end{figure}
These experiments so far may raise the concern of whether COOL networks are too restrictive to generalize well. Accordingly, the next set of experiments suggest three means to modulate generalization in COOL, namely:
\begin{itemize}
	\item changing the degree of overcompleteness,
	\item changing the depth of the architecture, and
	\item changing the softness parameter.
\end{itemize}
In short, by choosing the right hyperparameters one can tune the generalization ability of a model from pure discriminative networks on one end of the spectrum to behaving like a histogram approximation of the data distribution on the other side. This insight paves the way toward \emph{fitted learning}, i.e.~learning models with the right amount of generalization.

Figure~\ref{DOO} depicts the activation maps of the same concept in two trained networks with different degrees of overcompleteness (\emph{DOO}) on the two-circle problem. Both networks have the same architecture except for the output layer. These visualizations suggest that higher DOO can result in less overgeneralization. In other words, the less the DOO the more the network is likely to generalize over unseen regions. This observation is also supported by the intuition that a DOO of one is simply a traditional neural network.

\begin{figure}[ht]
	\begin{minipage}{0.495\columnwidth}
		\centering
		\centerline{\includegraphics[width=0.95\columnwidth]{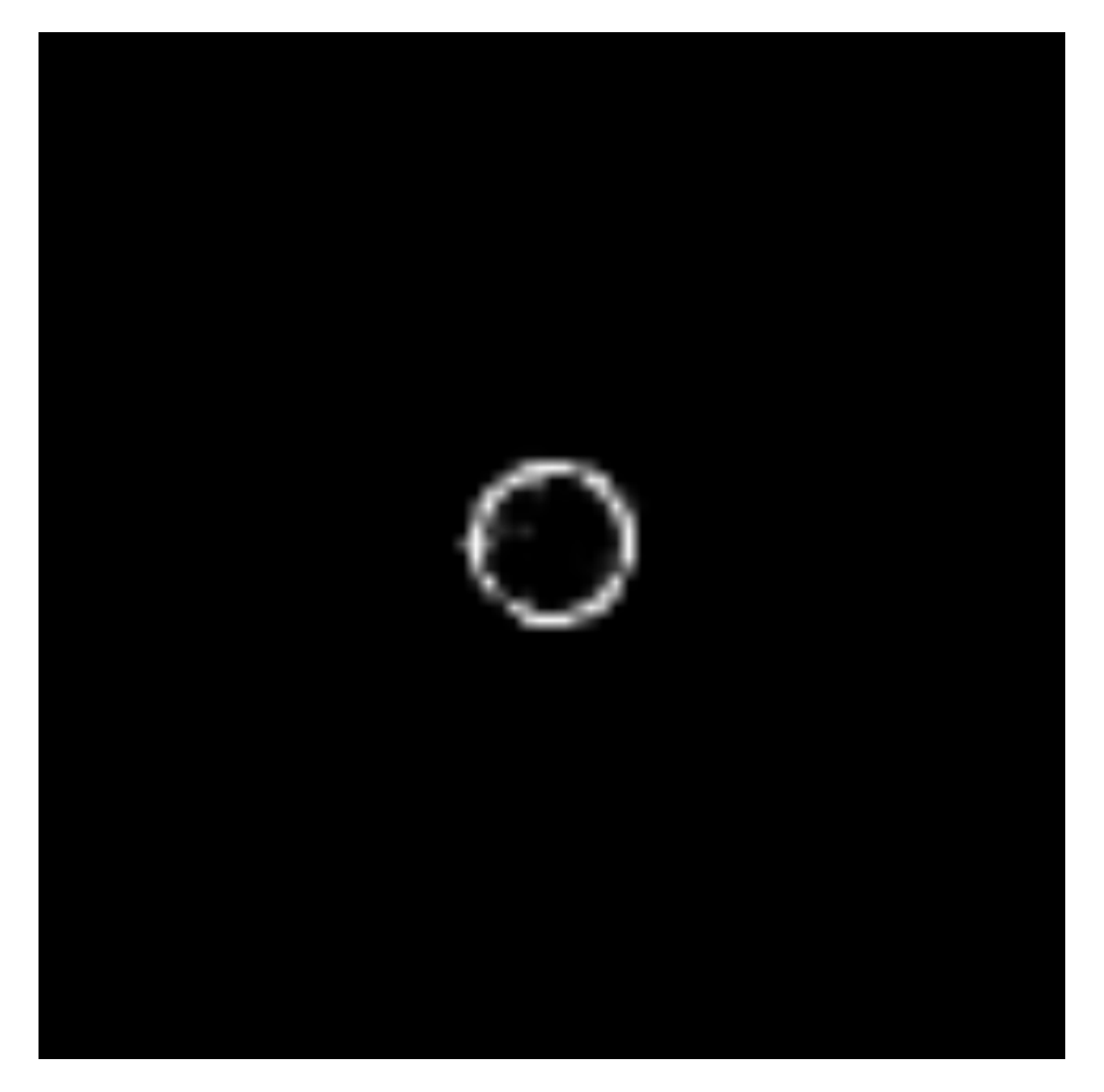}}
		\captionsetup{labelformat=empty,skip=8pt}
		\caption{(a)~DOO $= 10$}
	\end{minipage}%
	\begin{minipage}{0.5\columnwidth}
		\centering
		\centerline{\includegraphics[width=0.95\columnwidth]{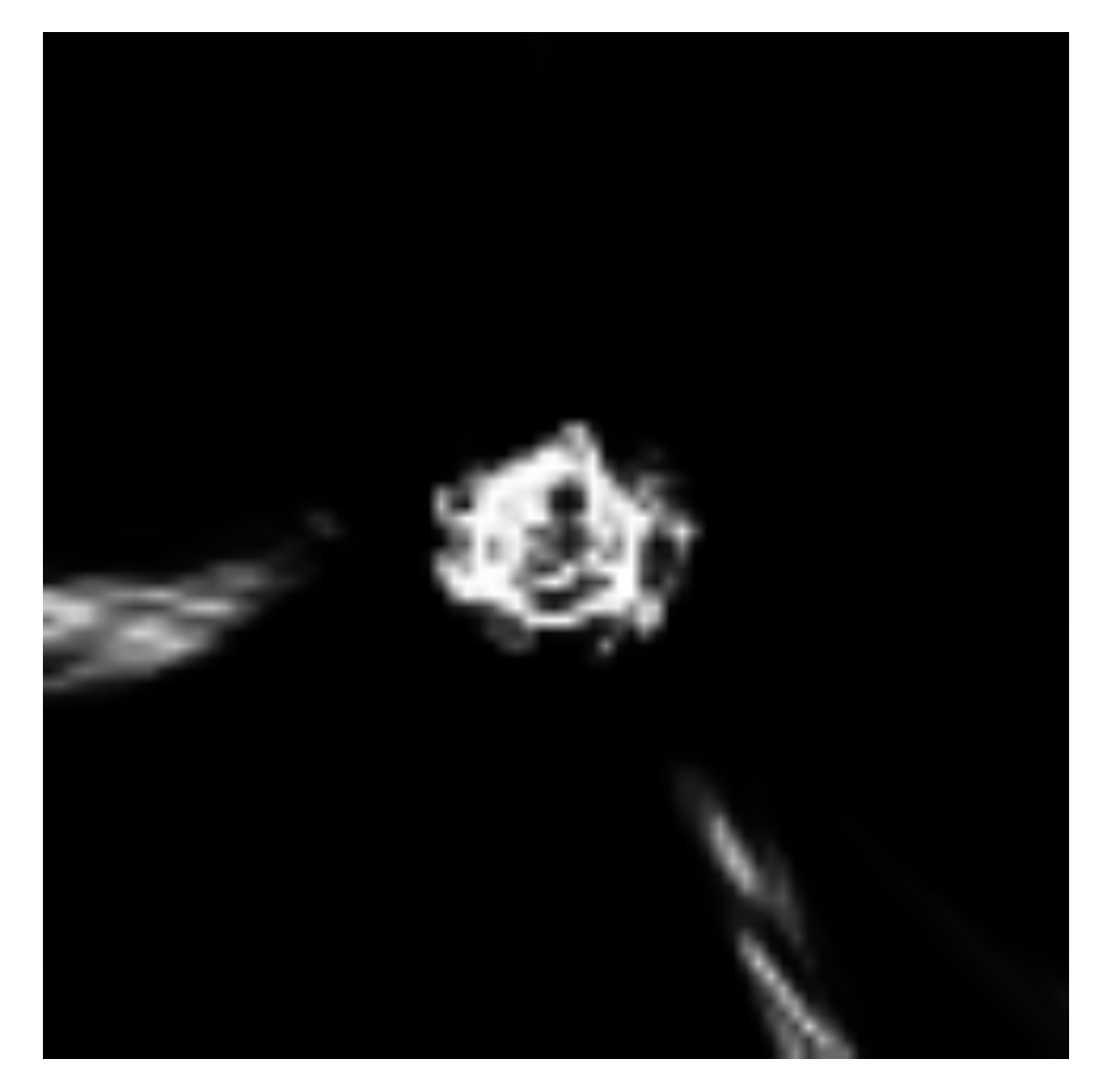}}
		\captionsetup{labelformat=empty,skip=8pt}
		\caption{(b)~DOO $=2$}
	\end{minipage}
	\setcounter{figure}{4}
	\caption{\textbf{Effect of changing DOO on the generalization ability of COOL.} Both plots show the activation map of the aggregate assigned to learn the inner circle.}
	\label{DOO}
\end{figure}

Next, two COOL networks with different depth but the same output layer (with DOO $= 5$) are compared. Figure~\ref{goingdeep} shows how adding more hidden layers can lead to more generalization ability but a higher risk of overgeneralization. In fact, in this experiment sometimes deep architectures lead to generalization over unwanted regions (e.g.~partially filling the circle shown in figure~\ref{goingdeep}).

\begin{figure}[ht]
	\begin{minipage}{0.5\columnwidth}
		\centering
		\centerline{\includegraphics[width=0.95\columnwidth]{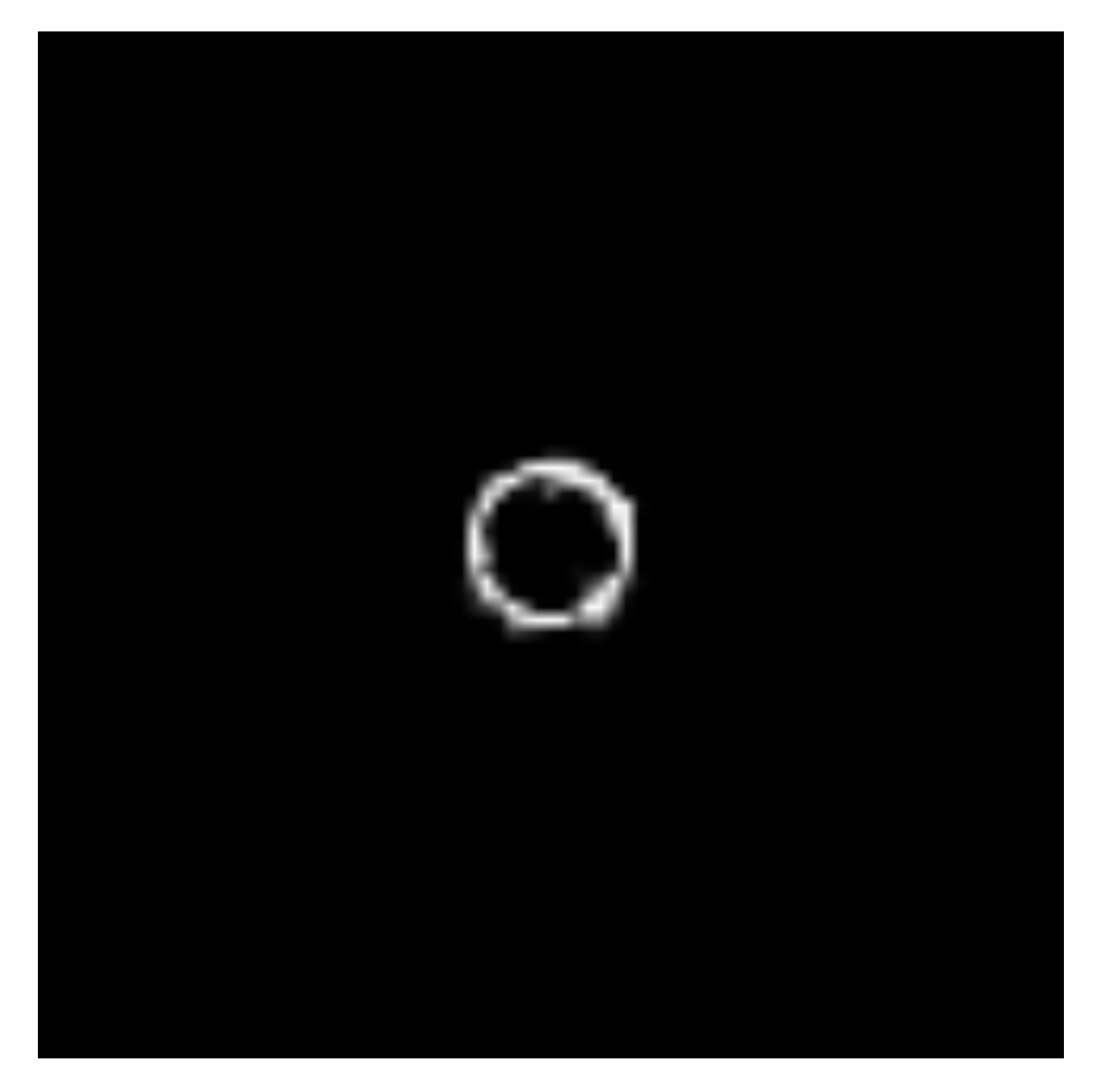}}
		\captionsetup{labelformat=empty,skip=8pt}
		\caption{(a)~1-layer COOL}
	\end{minipage}%
	\begin{minipage}{0.495\columnwidth}
		\centering
		\centerline{\includegraphics[width=0.95\columnwidth]{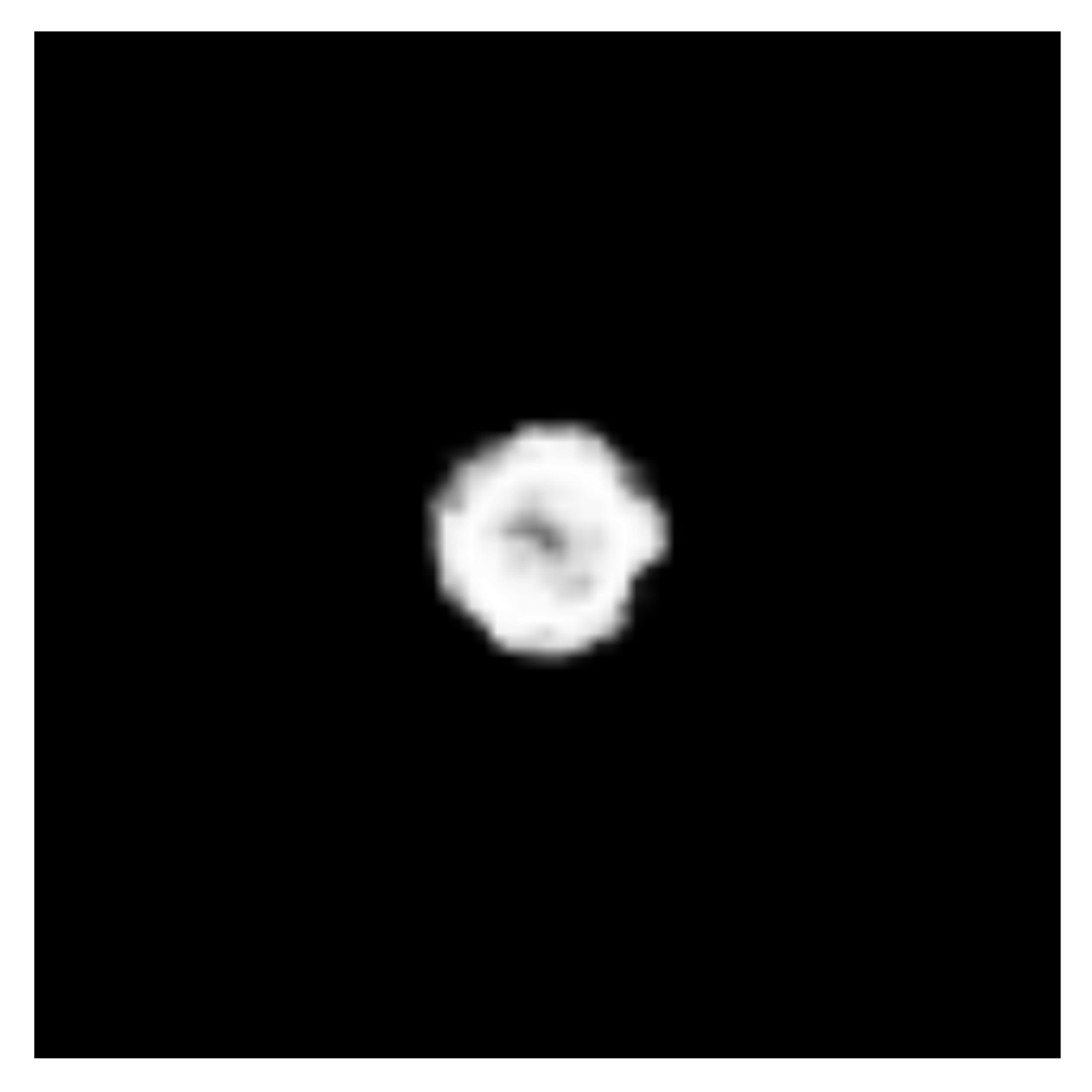}}
		\captionsetup{labelformat=empty,skip=8pt}
		\caption{(b)~5-layer COOL}
	\end{minipage}
	\setcounter{figure}{5}
	\caption{\textbf{Effect of applying deeper architectures on the generalization ability of COOL.} The activation maps associated with the inner circle in COOL networks with $1$ (a) and $5$ (b) hidden layers are shown, respectively. }
	\label{goingdeep}
\end{figure}

Finally, a brief experiment with the Iris dataset \citep{irisds} provides the opportunity to show the behavior of COOL in the presence of \emph{sparse} datasets. In general, sparse datasets, as opposed to \emph{dense} (or redundant) datasets, are those with few training instances per class such that even excluding a few examples can lead to significant negative impact on training. While deep learning techniques are mainly effective on dense datasets, sparse datasets are still important because in many domains training instances are scarce.

For visualization purposes, in this experiment only the first two input features of the Iris dataset are included in the training set. Figure~\ref{iris} depicts the dataset in this reduced space of features whereas the activation maps of output units of COOL networks assigned to three different types of iris are shown in figure~\ref{softness}. These plots show how changing softness parameter can shrink/extend the activation regions of each class, which can lead to a significantly better generalization.

Although in this paper we always set softness to $1.0$, as a rule of thumb it is more natural to choose softness so that the average accuracy of a model reflects its average activation value. The model's expected accuracy and activation can be estimated on a separate validation set, based on which the softness can be adjusted.  For example, if the average activation on the validation set is $0.9$ but the accuracy of the model is just $0.7$ ($70\%$), then if we set softness to $3.385$ it can calibrate the discrepancy between accuracy and average activation ($0.9^{3.385}\simeq 0.7$).  

In conclusion, the experiments in this section provide intuitive empirical evidence of the ability of COOL to learn the underlying data distribution of training instances while retaining the usual discriminative ability.  

\begin{figure}[ht]
	\vskip 0.1in
	\begin{center}
		\centerline{\includegraphics[width=\columnwidth]{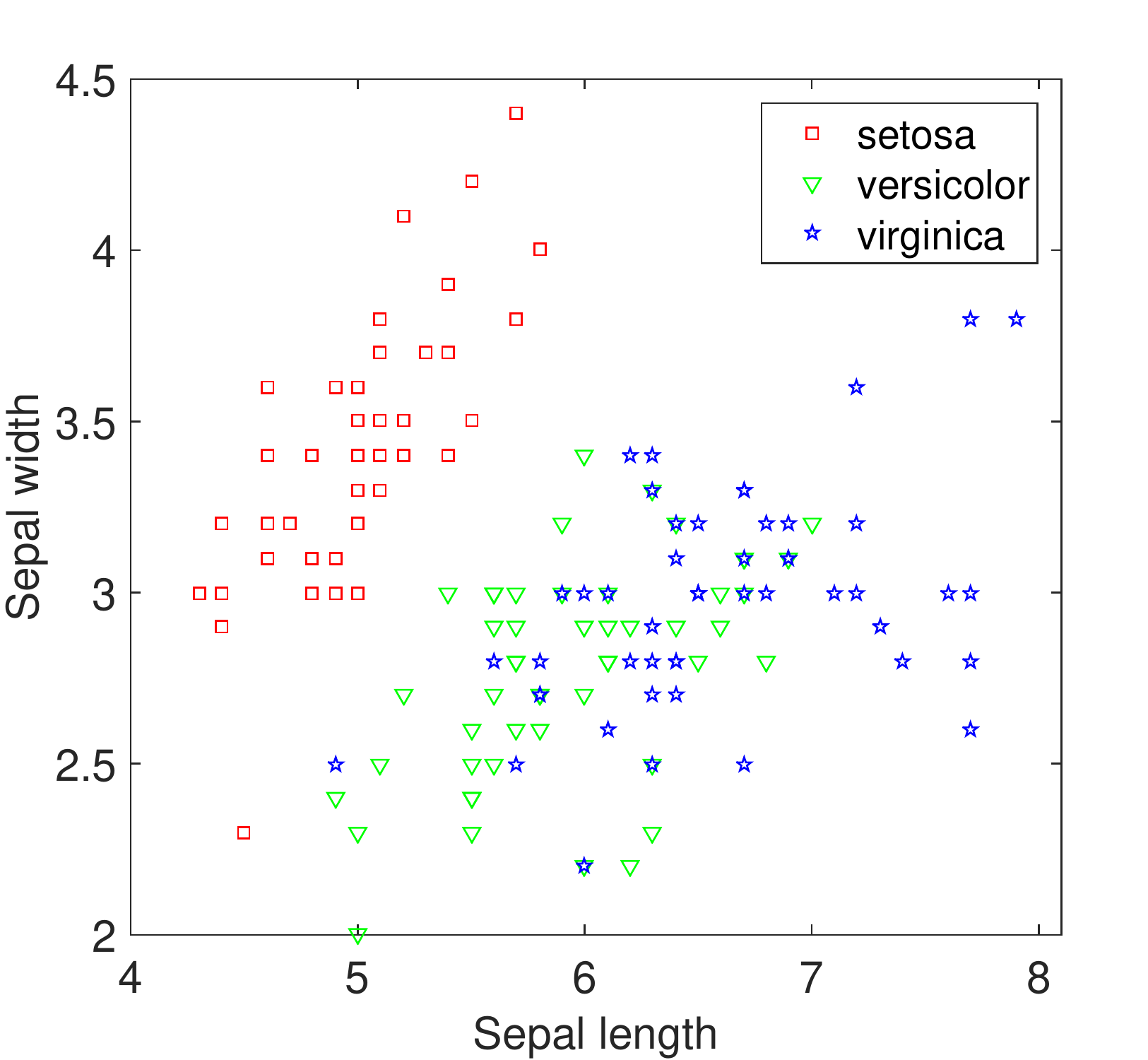}}
		\caption{\textbf{Iris dataset as an example of sparse dataset.} Note that the reduced input dimensionality in this problem makes it much harder than the original Iris problem.}
		\label{iris}
	\end{center}
	\vskip -0.1in
\end{figure}

\begin{figure}[ht]
	\begin{minipage}{0.33\columnwidth}
		\centerline{\includegraphics[width=0.999\columnwidth]{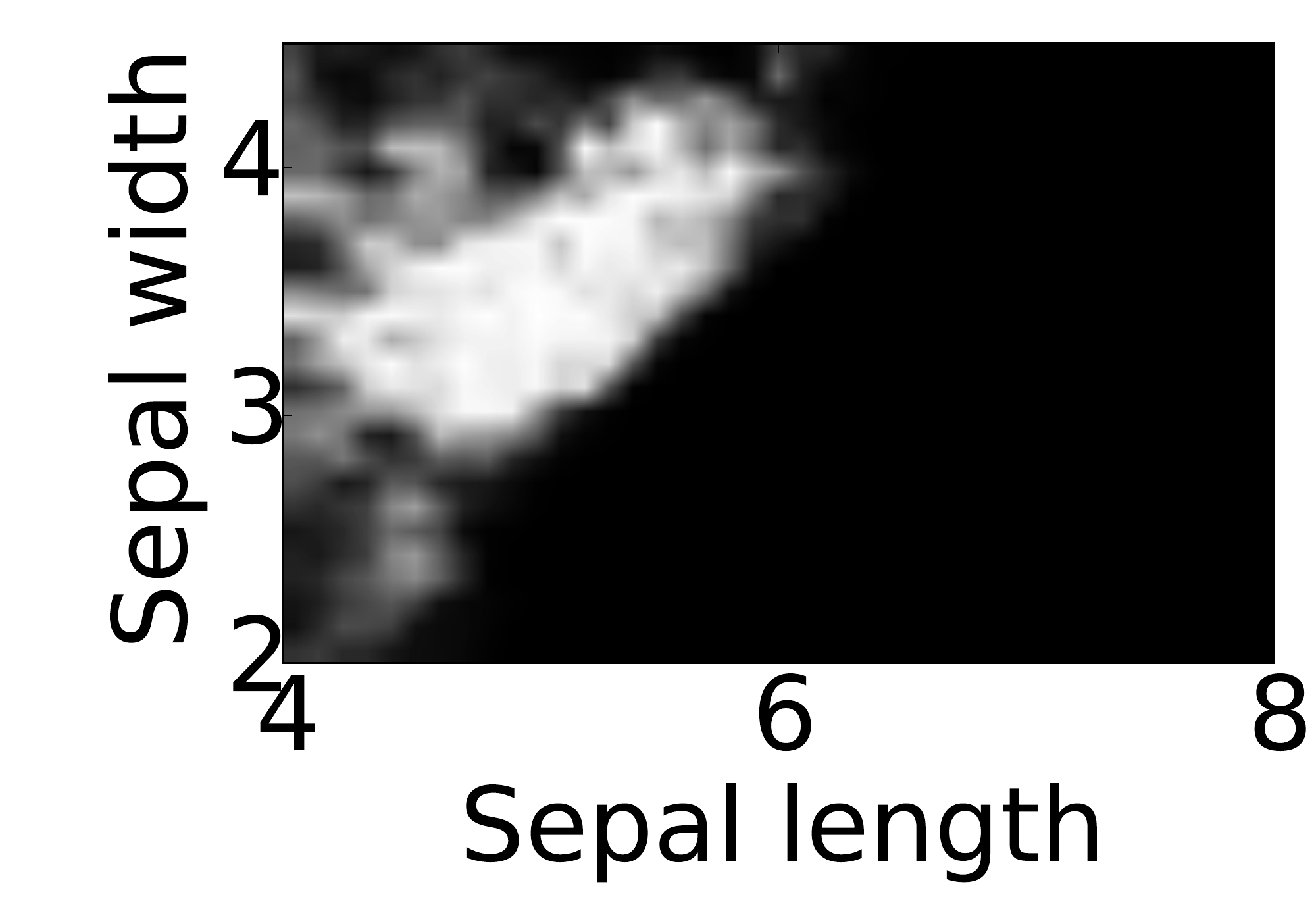}}
		\captionsetup{labelformat=empty,skip=2pt}
		\caption{ }
	\end{minipage}%
	\begin{minipage}{0.33\columnwidth}
		\centerline{\includegraphics[width=0.999\columnwidth]{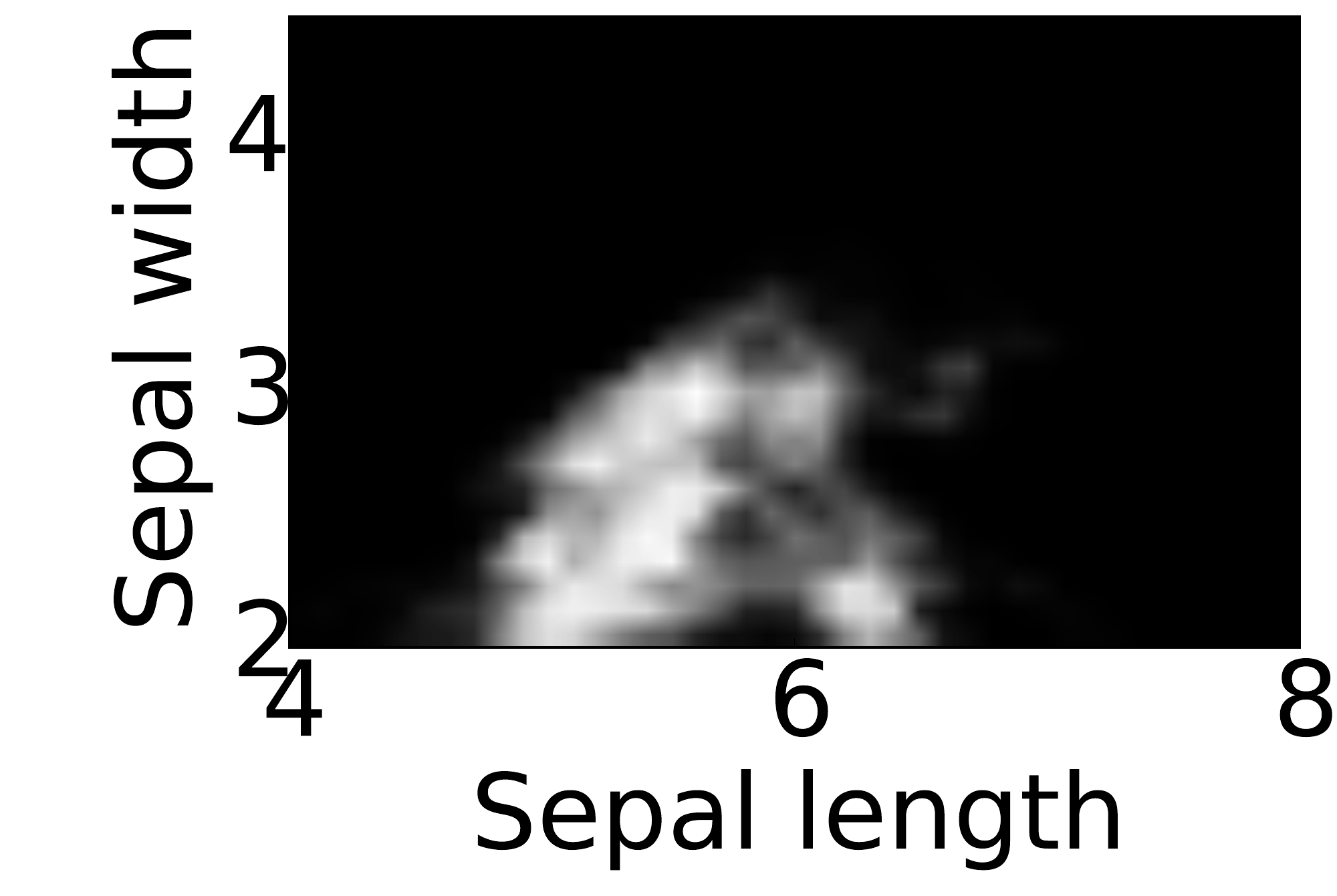}}
		\captionsetup{labelformat=empty,skip=2pt}
		\caption{(a)~Softness $0.5$}
	\end{minipage}%
	\begin{minipage}{0.33\columnwidth}
		\centerline{\includegraphics[width=0.999\columnwidth]{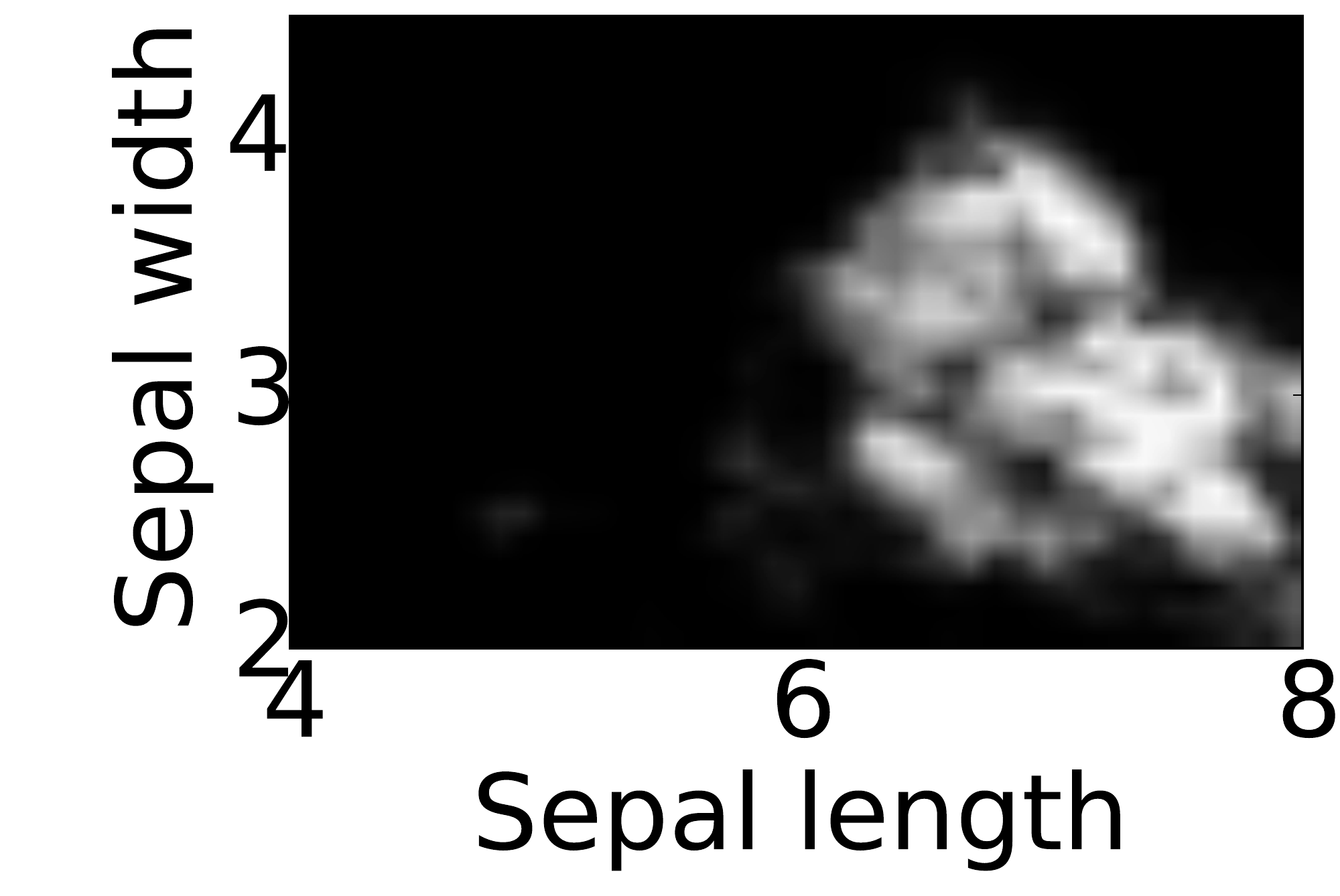}}
		\captionsetup{labelformat=empty,skip=2pt}
		\caption{ }
	\end{minipage}
	
	\begin{minipage}{0.33\columnwidth}
		\centerline{\includegraphics[width=0.999\columnwidth]{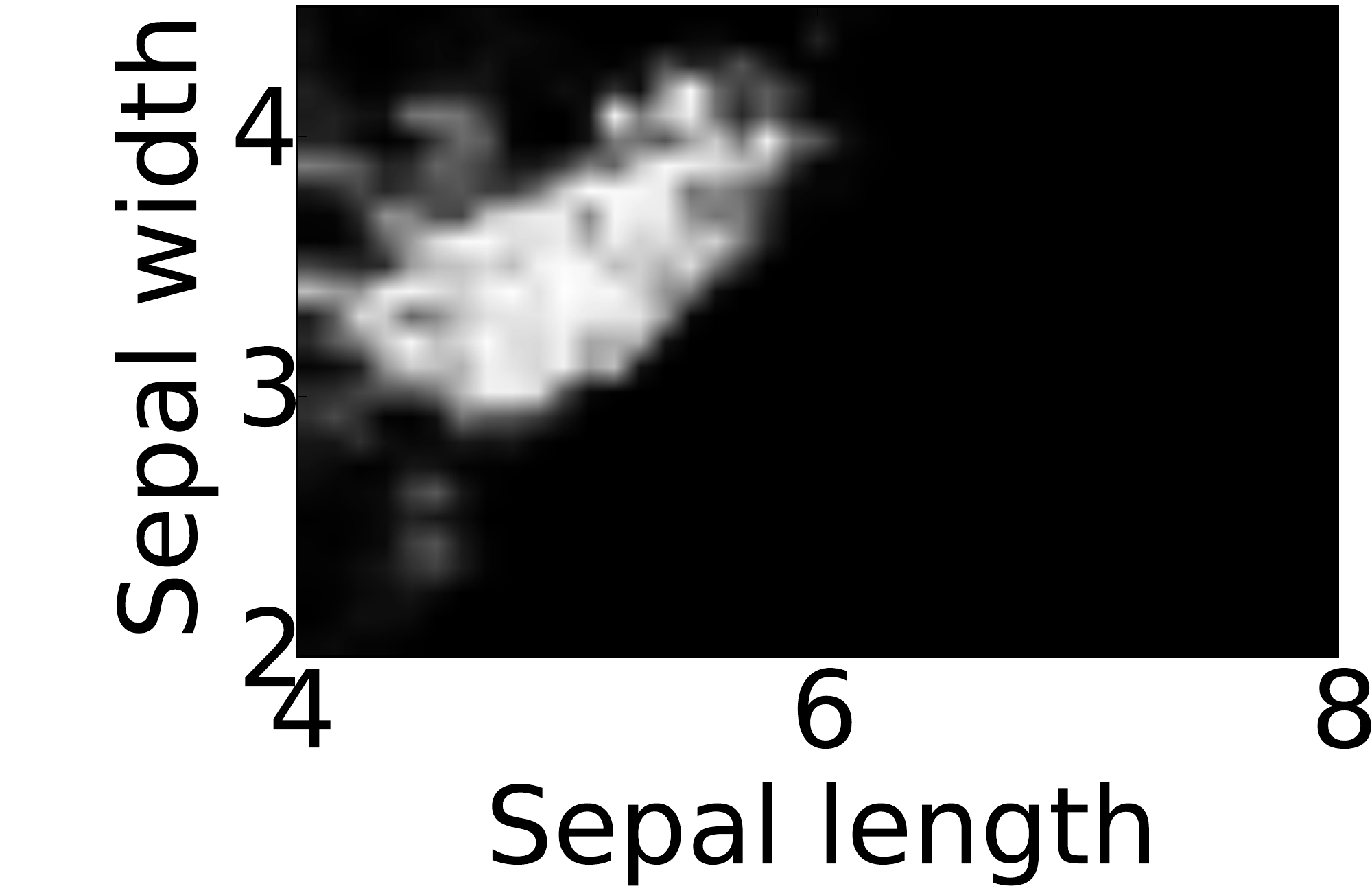}}
		\captionsetup{labelformat=empty,skip=2pt}
		\caption{ }
	\end{minipage}%
	\begin{minipage}{0.33\columnwidth}
		\centerline{\includegraphics[width=0.999\columnwidth]{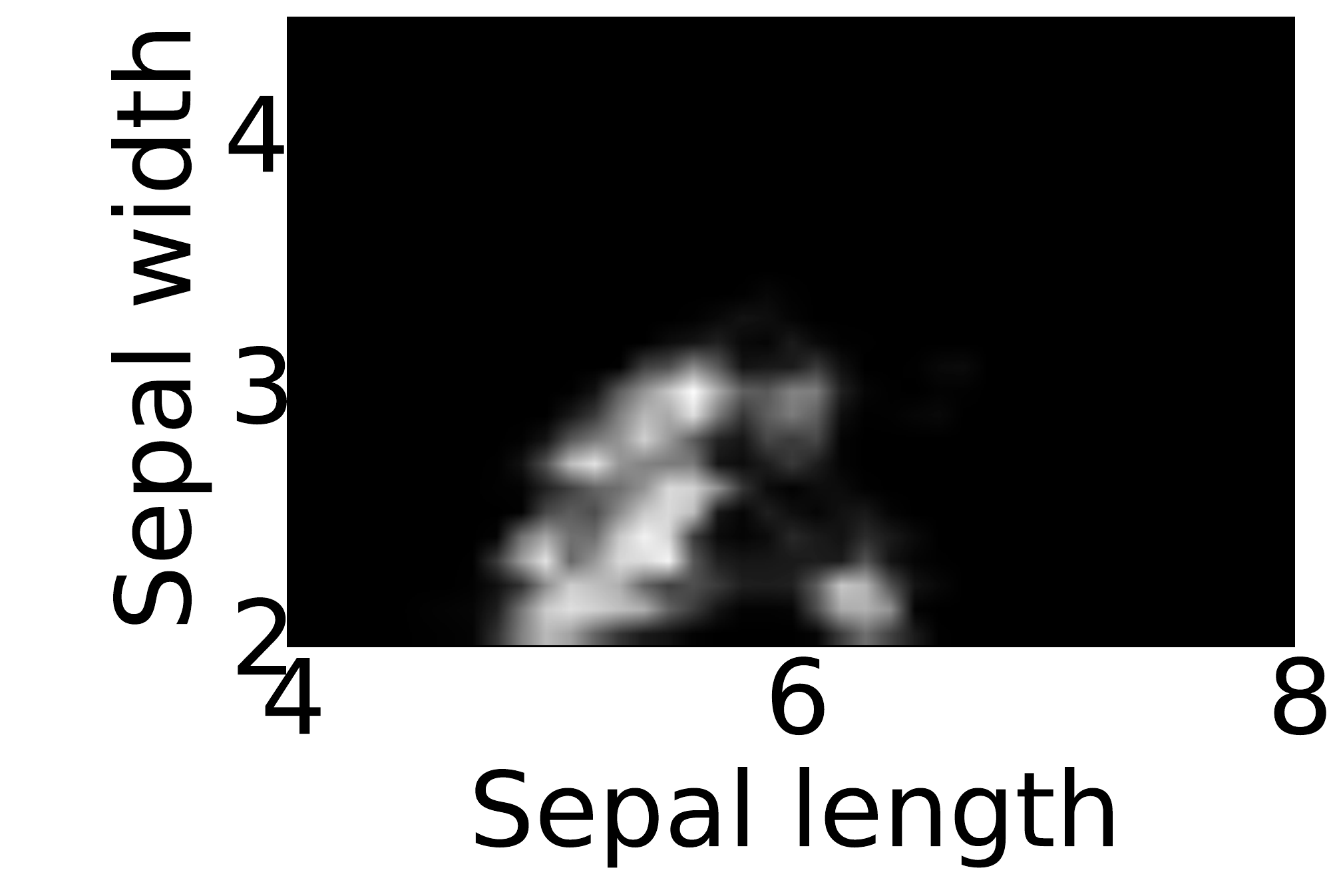}}
		\captionsetup{labelformat=empty,skip=2pt}
		\caption{(b)~Softness $1.0$}
	\end{minipage}%
	\begin{minipage}{0.33\columnwidth}
		\centerline{\includegraphics[width=0.999\columnwidth]{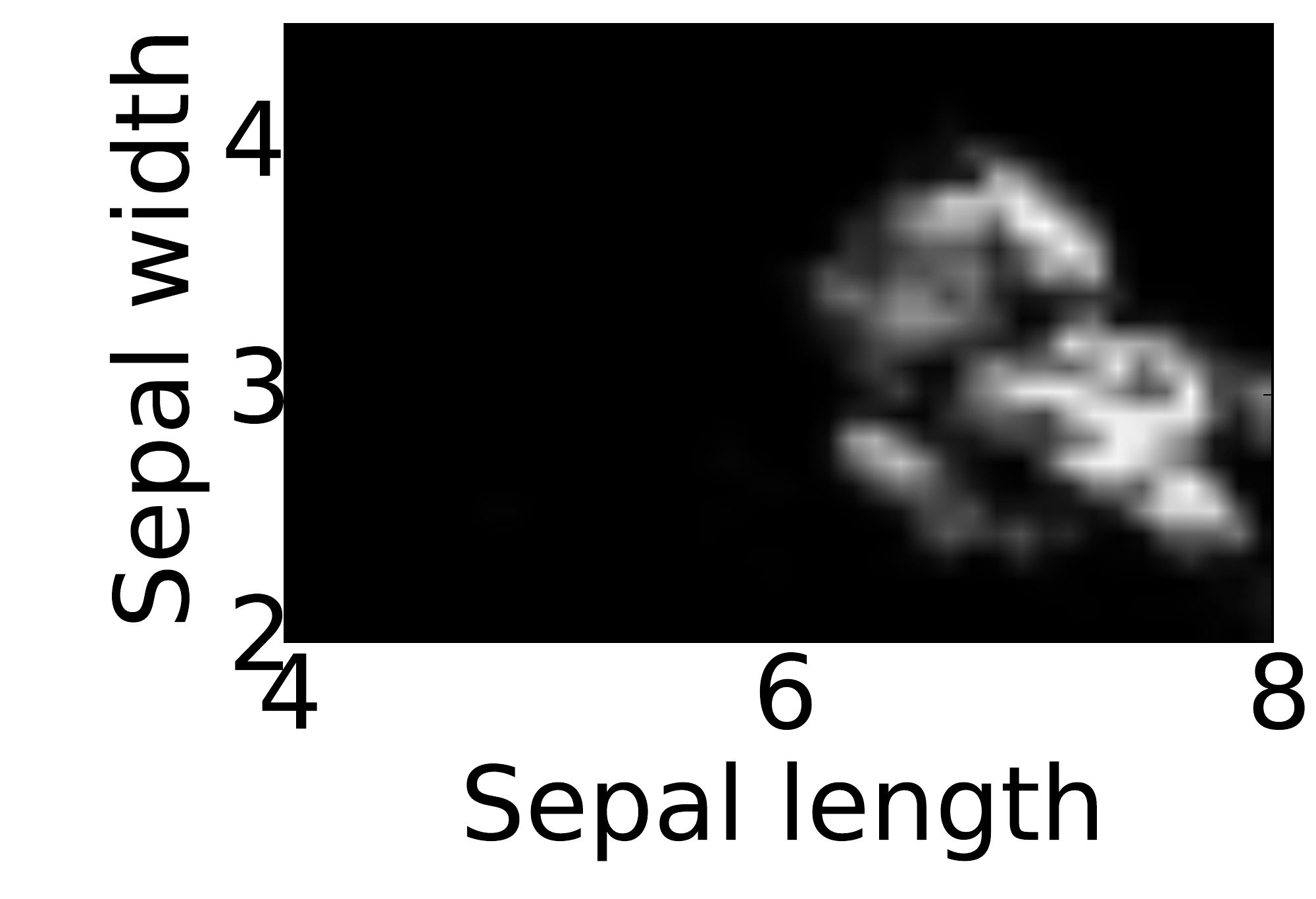}}
		\captionsetup{labelformat=empty,skip=2pt}
		\caption{ }
	\end{minipage}
	\setcounter{figure}{7}
	\caption{\textbf{How generalization ability can be tailored in COOL by changing the softness parameter.} The top three pictures (a) depict the activation map of three different classes  in a COOL network when the softness parameter is $0.5$. The bottom pictures (b) depict the activation maps in the same network when the softness parameter is $1.0$.}
	\label{softness}
\end{figure}
\section{Why the COOL Mechanism Works}
\label{whyworks}
This section explains how the two components of the COOL setup, i.e.~overcompleteness and competition, can work together against overgeneralization. In particular, it turns out that the overcomplete architecture leads to simultaneous training of an exponential number of models together while competition preserves diversity in the member units of a neuron aggregate that are trying to learn the same concept. The argument begins with an aggregate $X$ with $n$ member units, $X=\{x_1,x_2,...,x_n\}$.
\subsection{Overcompleteness}
The idea behind the overcomplete layer is that it produces a dynamic akin to training an exponential number of models all at once, which echoes the motivation behind dropout \citep{dropout2}. Assume that the output of each member unit is scaled just before multiplication in the test phase, i.e.~the output of member unit $x_i$ $\forall i\in\{1,2,...,n\}$ is transformed according to $S(x)=$max$(\omega \times x,1)$, where $\omega$ is the DOO.

In this realization each member unit's output can be interpreted as a probability value. Furthermore, any subset of an aggregate, except for the empty set, could in principle replace the original aggregate as the decision-maker. Consequently, training an aggregate is hard because it is analogous to training of $2^{n}-1$ neuron sub-aggregates with the same activation behavior at the same time. This challenge is further exacerbated by having an aggregate for each class in COOL.

\subsection{Competition}
Competition among the member units of the same aggregate is the instrumental mechanism of COOL that yields its protection against overgeneralization.
Recall that during the training of a particular instance there is only one \emph{active aggregate}, i.e.~the aggregate whose member units are trained for non-zero values. Because all the member units in the active aggregate are trained to learn the same function, one might expect them to converge to the same weights for each neuron in the aggregate. Interestingly, the competition induced by softmax prevents such an outcome.

To elaborate, assume $X$ is the current active aggregate for the input instance, $\chi$; $X$ has different member units and in practice in gradient descent there is at least one non-zero gradient\footnote{By "gradient" we mean the gradient of cost function with respect to the member units.} amongst $x_i$ $\forall i\in\{1,2,...,n\}$. The softmax function then implies that within a neighborhood $N$ of $\chi$, $\xi\in N(\chi)$,
\begin{equation}\label{softmax}
x_1(\xi)+x_2(\xi)+...+x_n(\xi) \simeq 1,
\end{equation}
where $x_i(z)$ is the activation of $x_i$ for the input instance $z$. Note that if all the member units are similar to each other, it is likely that they are all within a close proximity in the space of the cost function. This closeness stems from all member units applying the same cost function, having the same input, and targeting the same value. Therefore, in a neighborhood N of $\chi$ equation~\ref{softmax} yields:
\begin{equation}
\frac{\partial E}{\partial x_n} =\frac{\partial E}{\partial x_1} \frac{\partial x_1}{\partial x_n}\simeq -\frac{\partial E}{\partial x_1},
\end{equation}
where $E$ is the cost function. In other words, the gradient of $x_n$ is \emph{competing} (i.e.~pulling in a different direction) with the gradient of $x_1$. Note that in this discussion, $x_n$ is assumed to be the redundant parameter of softmax, which is a function of other independent parameters $x_1,x_2,..,x_{n-1}$.

As a result, converging to the same unit is not attractive to the member units when starting from different initial parameters.

In short, assigning several competing units to the same concept can be thought of as a \emph{game of changing model parameters} between member units of the active aggregate that can lead to different partitioning of the input space among them. Therefore we expect these units to preserve their diversity for arbitrary inputs far from the training instances and only to maintain consensus over the regions proximal to the training set. To further illustrate this principle, figure~\ref{nocompetition} depicts the activation map of a COOL network that lacks the competition component. This neural network is trained on the two-circle problem from the visualization experiments. Compared to COOL (figure~\ref{rings}), figure~\ref{nocompetition} shows that without competition all the member units associated with the same concept can easily converge (and hence overgeneralize). Finally, note that the two plots in figure~\ref{nocompetition} are not the exact complement of each other, which indicates that even without competition there is still some advantage over conventional neural networks.
As a final note, it is important to highlight that \emph{while competition encourages diversity amongst output units during training, the weight initialization scheme determines the amount of diversity at the very beginning}. One way to encourage such initial diversity is by applying higher variance to all the model parameters, e.g.\ through a standard Gaussian distribution. However, we empirically found that initializing the parameters of the output layer with a high variance suffices for this purpose.  

\begin{figure}[ht]
	\begin{minipage}{0.5\columnwidth}
		\centering
		\centerline{\includegraphics[width=0.95\columnwidth]{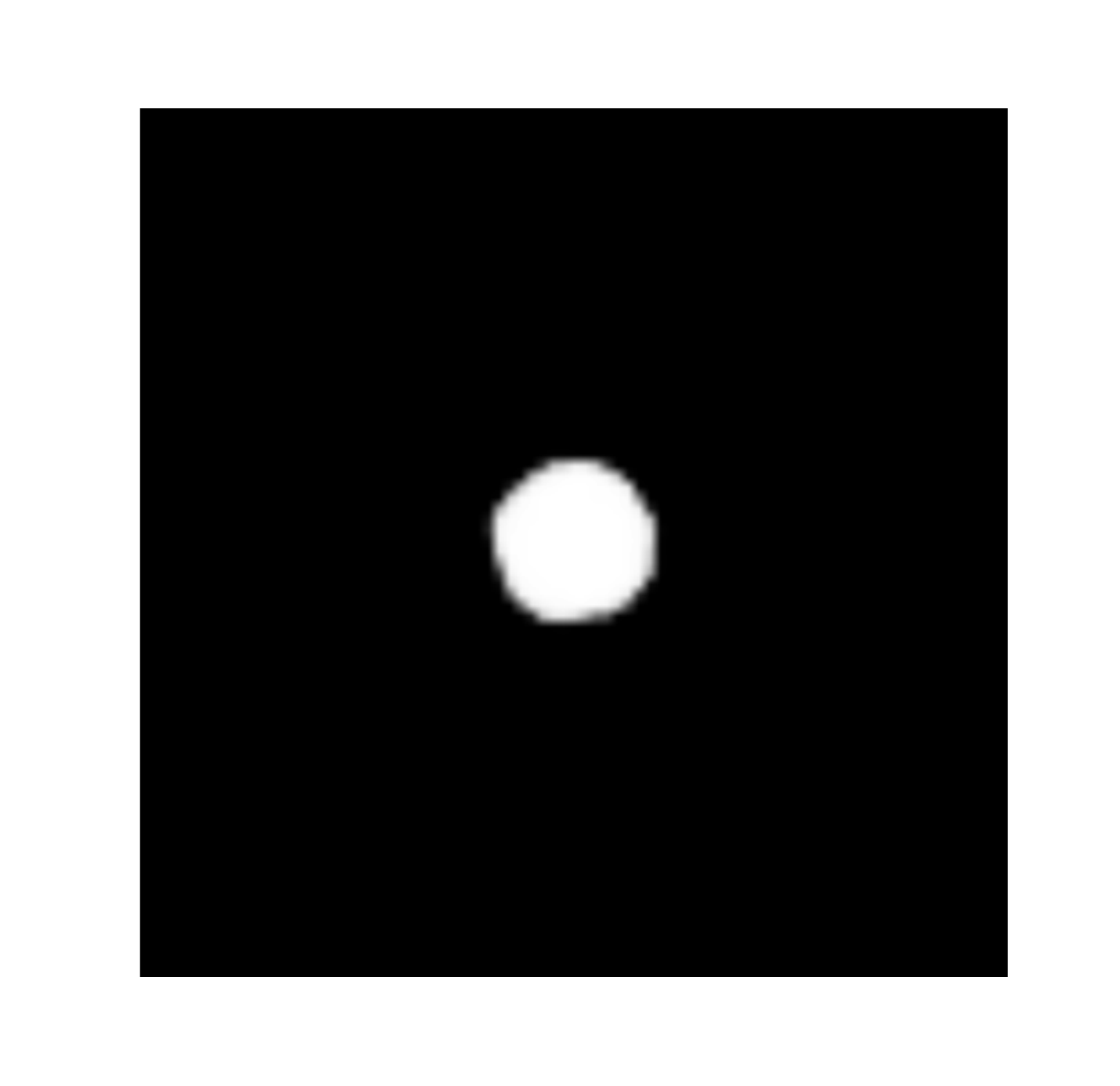}}
		\captionsetup{labelformat=empty,skip=0pt}
		\caption{(a)~Inner circle output}
	\end{minipage}%
	\begin{minipage}{0.5\columnwidth}
		\centering
		\centerline{\includegraphics[width=0.95\columnwidth]{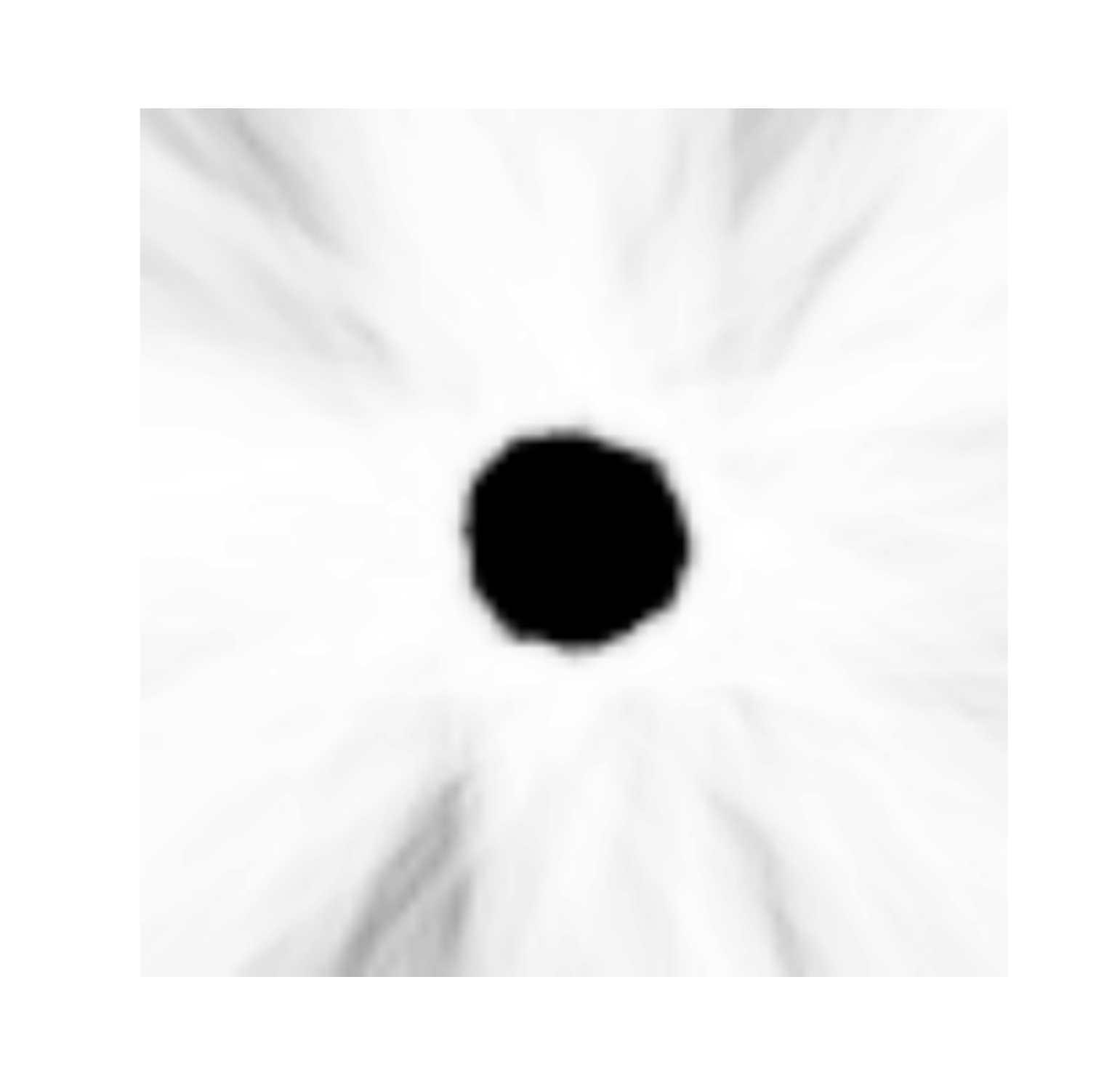}}
		\captionsetup{labelformat=empty,skip=0pt}
		\caption{(b)~Outer circle output}
	\end{minipage}
	\setcounter{figure}{8}
	\caption{\textbf{The effect of removing competition from COOL.} These plots depict the activation map of aggregates corresponding to (a) inner and (b) outer circles in the two-circle problem when softmax is removed (compare to figure \ref{rings}). }
	\label{nocompetition}
\end{figure}

\subsubsection{Non-uniform Distribution Problem}
In the next visualization experiment, a more challenging data distribution, namely a logarithmic spiral enclosed within a circle (Figure \ref{spiralds}), is applied to evaluate how well different models can capture non-uniform densities. We also trained a \emph{separate GAN for each class} to show how well GAN discriminators performs in rejecting out-of-distribution examples. 
Here the logarithmic spiral is $4\pi$ long and the number of training samples is proportional to the polar angle and not the curve length. Therefore the points closer to the center of the curve tend to have more training samples (higher density) and the density is reduced towards the end of the curve. 

\begin{figure}[t]
	\vskip 0.1in
	\begin{center}
		\centerline{\includegraphics[width=\columnwidth]{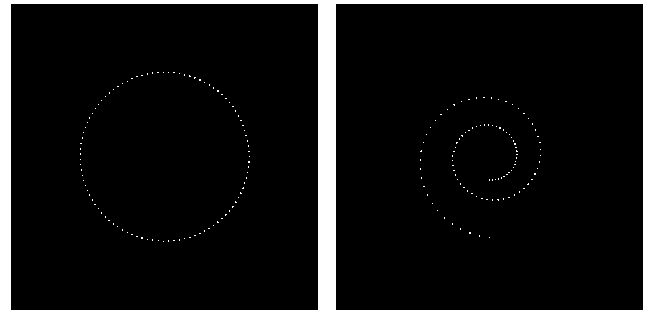}}
		\caption{\textbf{Spiral-based non-uniform training set.} Left: outer circle representing one of the classes. Right: logarithmic spiral representing the other class. Note the change of sampling rate in the spiral showing a non-uniform distribution.}
		\label{spiralds}
	\end{center}
	\vskip -0.1in
\end{figure}

The non-uniform distribution experiment applies the same setup as the two-circle problem with the exception that the number of training epochs  for COOL network is increased to $5$,$000$ in this one. 
The training set is again created by sampling $500$ points from both a circle centered at the origin with radius $1.1$ and a logarithmic spiral curve:

$x(t)=ae^{bt}cos(t)$ and
$y(t)=ae^{bt}sin(t)$,

where $a=0.3$ and $b=0.1$.

Figure \ref{logarithmicspiral} shows COOL can again capture this (now non-uniform) density precisely. On the other hand GAN, though almost successful at capturing the shape of the distribution, was not able to capture the non-uniformity of the distribution in all of 20 attempts\footnote{GAN provides a powerful framework but needs a careful architecture design and training procedure. Here we applied the original version of GAN \cite{gan}.}. This shortfall is probably related to the well-known mode collapse problem \citep{gantutorial} in GAN networks. The visualizations from this and the previous experiment of simple two-dimensional domains give an intuitive sense of the unique behavior of COOL.
  
\begin{figure}[!t]
\begin{minipage}{0.32\columnwidth}
		\centering
		\centerline{\includegraphics[width=0.95\columnwidth]{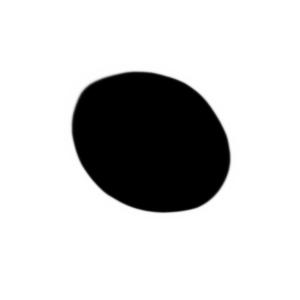}}
		\captionsetup{justification=centering,labelformat=empty,skip=0pt}
		\caption{(a)~Regular NN, circle}
	\end{minipage}
	\begin{minipage}{0.3\columnwidth}
		\centering
		\centerline{\includegraphics[width=0.95\columnwidth]{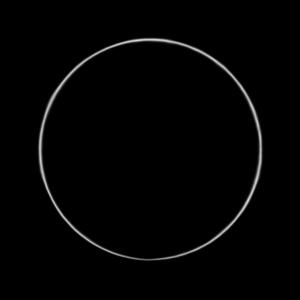}}
		\captionsetup{justification=centering,labelformat=empty,skip=0pt}
		\caption{(b)~COOL NN, circle}
	\end{minipage}%
\begin{minipage}{0.3\columnwidth}
		\centering
		\centerline{\includegraphics[width=0.95\columnwidth]{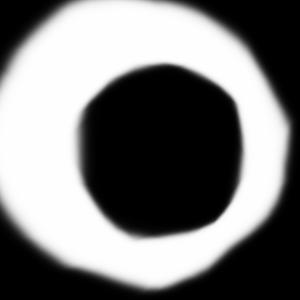}}
		\captionsetup{justification=centering,labelformat=empty,skip=0pt}
		\caption{(c)~GAN disc., circle}
	\end{minipage}%

\hspace{1mm}
\begin{minipage}{0.3\columnwidth}
		\centering
		\centerline{\includegraphics[width=0.95\columnwidth]{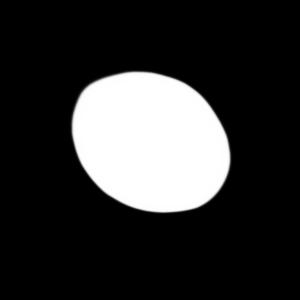}}
		\captionsetup{justification=centering,labelformat=empty,skip=0pt}
		\caption{(d)~Regular NN, spiral}
	\end{minipage}%
\begin{minipage}{0.3\columnwidth}
		\centering
		\centerline{\includegraphics[width=0.95\columnwidth]{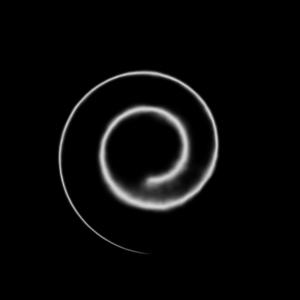}}
		\captionsetup{justification=centering,labelformat=empty,skip=0pt}
	\caption{ (e)~COOL NN, spiral}
	\end{minipage}
    	\begin{minipage}{0.3\columnwidth}
		\centering
		\centerline{\includegraphics[width=0.95\columnwidth]{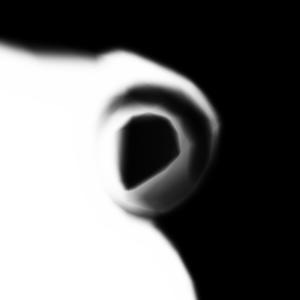}}
		\captionsetup{justification=centering,labelformat=empty,skip=0pt}
		\caption{(f)~GAN disc., spiral}
	\end{minipage}%
    \setcounter{figure}{10}
	\caption{\textbf{The ability of different models to capture non-uniform distributions of data}. In this problem a logarithmic spiral (class B), second row, is enclosed within a circle (class A), first row. The plots show how a regular discriminator model (a,d) only finds a border between the two classes, while COOL (b,e) can capture the data distribution as well as learn the classification task. A GAN (c,f) is (partially) able to capture the distribution but is not able to learn the classification task (note the overlap between the activation maps of the output units for class A and B in GAN).
    }
	\label{logarithmicspiral}
\end{figure}

\section{The Fooling Problem}
A phenomenon called \emph{fooling} has recently attracted attention in the deep learning community \citep{szegedy,nguyen}.  The main observation is that neural networks can be tricked into outputting with confidence the wrong class for instances well outside their training distribution. In other words, the fooling problem is a direct consequence  of overgeneralization, which is more prominent in deep architectures. 

To clarify the fooling problem we begin with a simple example:
The classification task is to determine whether a fruit is a watermelon or an apple based on the weight of the fruit in kilograms. A possible training set is then $T=\{(0.25,apple),(0.2,apple),(5.25,watermelon),$ $(7.3,watermelon)\}$. In this example a discriminative model, such as a SVM, will find a decision boundary that can be translated into $if ~weight>2.75~ then~ watermellon,~ otherwise~ apple$. Often discriminative models can output probability values based on the underlying decision boundary as well \citep{duda}. For example, in such a model if an input is close to $2.75$, say $2.5$, it would be classified as an apple but with a lower probability than a fruit with weight $1.5$. The fooling problem occurs when the input is not drawn from the underlying data distribution, e.g.~if we try a fruit that weighs 50 kg most discriminative models trained as above would classify it as a watermelon even though such a fruit possibly does not exist.

It is worth noting that ensemble techniques on their own have difficulty in mitigating fooling problem \citep{szegedy} because they merely generate a better decision boundary, in the form of a larger margin. However, fooling is not a problem of the margin.

The biggest downside of fooling is the fact that it is \emph{dataset-dependent} and not model-dependent. In other words, one can easily find fooling (false positive) examples that can fool almost \emph{all} discriminative models only based on a limited knowledge of the training set \citep{szegedy,nguyen}.

This section first reviews reasons behind fooling. Then it provides a simple way to measure how easily a trained neural network can be fooled based on a generative adversarial approach. Finally, it compares the fooling in COOL versus conventional neural networks on the MNIST dataset. The results suggest that the COOL mechanism is capable of preventing fooling in neural networks.
\subsection{Reasons Behind Fooling}

\citet{goodfellow} postulate a linear explanation for the existence of adversarial examples, where small perturbations to an existing training example can lead to significant changes in the output in a high-dimensional problem. They also extend this explanation in their appendix to the more general case of an arbitrary fooling (false positive) example, i.e.\ one not necessarily similar to a training example, such as in \citet{nguyen}. This paper (and COOL) focuses on the latter case.

In general, suppose that training instances are sampled from an unknown probability distribution $D$. Consider the classification of an arbitrary input example $x$ by neural network $N$; $x$ is a fooling (false positive) example if it is classified with a high confidence by the network $N$ while \emph{not} being drawn from the probability distribution $D$. A significant factor behind fooling is then that discriminative classifiers tend to overgeneralize. That is, they classify with confidence many input instances regardless of their resemblance to the training data. Several active output units for a single instance implies low confidence of the classifier, but a single dominant output unit for an outlier example implies fooling the model. Because these kinds of learning models implicitly draw borders in the instance space, the distance between input points in the same region is in effect ignored, resulting in the same degree of classification confidence for input instances that can be far away. Interestingly, COOL networks inherently address this issue.      

\subsection{Generating Fooling Instances}

Previous methods for generating fooling (false positive) images from \citet{szegedy} and \citet{nguyen} search directly for images that fool the network. In contrast, here we introduce a third option that does not require solving a constrained optimization problem: A random input instance $x$ is fed into a new trainable neural network $g$, called the \emph{fooling generator network} (FGN), whose output is passed to the actual model $f$. In other words, $g(x)$ is the fooling input to the network instead of $x$. Gradient descent can train network $g$ such that $g(x)$ generates a good fooling (false positive) image. 

During this procedure, $x$ and parameters of $f$ are fixed and only $g$ is being trained. More formally, $f:\Re^n\rightarrow[0,1]^k$ is a mapping from input vectors to a probabilistic target vector space $\rho$, $h:\rho\rightarrow\{1,...,k\}$ is an invertible mapping from $\rho$ to a discrete set of labels, and $g:\Re^m\rightarrow \Re^n$ is a mapping from input vectors of size $m$ to vectors of size $n$ (note that $m$ can be equal to $n$). The key idea is thus to solve the optimization problem $Minimize ||f(g(z))-h^{-1}(l) ||_2$, for a given input $z\in \Re^m$ and target label $l\in\{1,...,k\}$, where $h^{-1}(.)$ denotes the inverse of $h(.)$. Note that $||.||_2$ can be replaced by cross-entropy in this optimization problem, which is the procedure in the experiments that follow.

This approach has several advantages:
First,	there is no need for \emph{constrained} optimization because one can control the outputs of network $g$ by choosing the right activation function. For example, a sigmoid function can give an output in the range $(0,1)$.
Second, a good choice of architecture can indirectly impose desirable constraints on the generated fooling (false positive) examples, e.g.~when dealing with images, a convolutional FGN $g$ imposes some natural image properties on the generated fooling (false positive) images.
Finally, almost all the components of this approach are already provided in most machine learning packages.

\subsection{MNIST Fooling Experiment}

In this experiment first a COOL network and a conventional CNN are trained on the MNIST dataset.
Then $20$ trials are attempted to generate fooling instances for each model. A trial consists of training a FGN activated with a random input to trick the model and is considered successful if the model classifies the generated fooling (false positive) image with more than $99\%$ confidence. However, if such a fooling (false positive) example is not found before $10,$$000$ parameter updates of the FGN then the trial is a failure. More details on this experiment are in the Appendix.

COOL preserves the generalization ability of CNNs for this task: the classification accuracies of COOL and conventional CNNs were $99.18\%$ and $99.14\%$, respectively. At the same time, overall, in $200$ trials ($20$ for each digit), the FGN approach fooled the conventional CNN with a $100\%$  success rate versus a $47\%$ fooling rate for the COOL CNN. Figure~\ref{generateddigits} depicts examples of typical generated fooling (false positive) images for both the conventional and the COOL CNN. Interestingly, on average a conventional CNN is fooled before $13$ updates of the FGN parameters whereas the COOL CNN needs more than $5$,$000$ updates, which suggests a drastic shrinkage of the high confidence classification regions within the COOL CNN.

Another interesting observation is that most often (more than $60\%$ of the time) the conventional CNN confidently classifies a random image (uniform random noise without any modification) as digit $8$ while COOL is \emph{never} fooled by a random image, which is further evidence of the severity of the fooling problem in conventional neural networks. Also, while the conventional CNN has a $100\%$ fooling rate for each individual digit, the FGN approach exhibits variable success for different digits with COOL. For example it failed to trick the COOL CNN for the digit $1$ in all of its attempts. This result suggests a more effective approach to preventing fooling may involve a specific DOO for each aggregate.

Finally, it is important to note that this experiment assumes complete access to the models that are meant to be fooled. In a real-world scenario, fooling is likely based on limited access to the training set or some input/output samples from the trained model. The low fooling rate of COOL networks in this experiment with complete access to the model therefore hints that they may in effect prevent fooling in many real scenarios.  
\begin{figure}[!t]
	\vskip 0.1in
	\begin{center}
		\centerline{\includegraphics[width=\columnwidth]{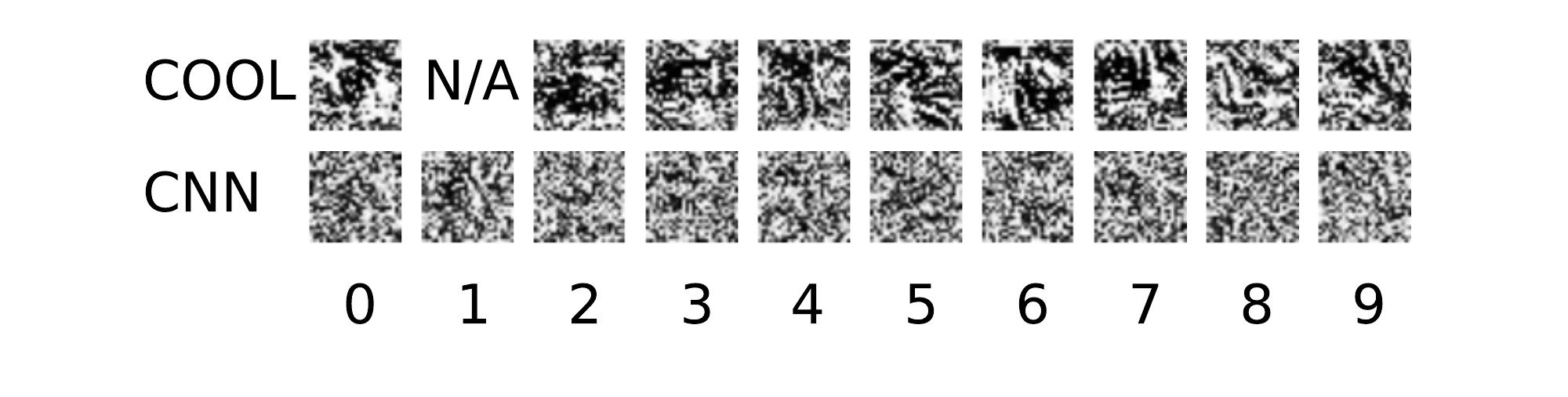}}
		\captionsetup{skip=0pt}
		\caption{\textbf{Example fooling images generated by the fooling generator network for the CNN.} Left to right, each image represents digits 0 to 9, respectively. These images are generated by a FGN consisting of a fully connected layer to produce plausible images. They are classified by the trained conventional CNN (bottom) and COOL CNN (top) with at least 99\% confidence. The top image for $1$ is blank because the FGN could not generate any fooling (false positive) image in all of its trials for this digit. Even though none of these images are close to real digits a qualitative difference between the images on top and bottom is evident.}
		\label{generateddigits}
	\end{center}
	\vskip -0.1in
\end{figure}
\section{Fitted Learning and Separable Concept Learning} \label{definitions}
So far COOL has been offered as a mechanism to prevent overgeneralization in neural networks. As a consequence COOL networks are able to prevent fooling or at least make it relatively difficult for an adversary. A natural question that follows is how to measure the amount of overgeneralization in a model. This section defines such a measure by introducing the \emph{generalized classification problem} that leads to a notion of \emph{extended empirical risk}, which is a natural extension of empirical risk \citep{vapnik}.

Recall from classical statistical learning theory \citep{vapnik}, where one assume access to a finite collection of i.i.d.\ samples in a training set, $T=\{(x_i,y_i)\}_{i=1}^{n} \sim D(X,Y)$, where $X$ is the input space of training set $T$, $Y$ is the output space, and $D$ is the data distribution over $X \times Y$. In the canonical classification problem, we also assume $Y$ is a finite set of objects of size $c$.

Given a non-negative, real-valued loss function $L(y',y)$ that measures the loss of predicting $y$ as $y'$, the risk $R$ of a function $f:X \rightarrow Y$, called a \emph{hypothesis}, is defined as: 
\begin{equation} \label{risk}
R(f)=E(L(f(x),y)).
\end{equation}
The goal of classification is to find a hypothesis $f^*$ such that $R(f^*)$ is minimal. Note that every hypothesis can be associated with an underlying interpretation of $P(Y|X)$. 
To avoid trivial solutions, hypotheses are usually restricted to a class of functions $F$. In other words:
\begin{equation} \label{optim}
f^*=\mathrm{argmin}_{f\in F} R(f).
\end{equation}
Solving equation \ref{optim} is theoretically impossible because of ignorance about  $D(X,Y)$; thus an approximation of $R(f)$ is expressed by $R_e(f)$ :
\begin{equation}
R_e(f)=\frac{1}{n}\sum_{i=1}^{n} L(f(x_i),y_i).
\end{equation}
Finding $f^*$ using the $R_e$ instead of $R$ is called \emph{empirical risk minimization}, which is a bedrock principle of statistical learning theory \citep{vapnik}.

\subsection{Introducing the Generalized Classification Problem}

In conventional classification, a learning model simply draws a border between different concepts. One drawback of this strategy, as discussed earlier, is an inherent overgeneralization. \emph{Generalized classification}, introduced here, addresses this issue by asking for the right amount of generalization. Moreover, in many real-world applications a comprehensive knowledge of the concepts is not available during training, or even worse, some new concepts might emerge over time. For example, in classification of a certain fish species we cannot rely on our current knowledge of fish because new species are discovered over time. Thus, ideally we want our models to adapt to new emerging concepts without significant retraining. 

Another aspect of conventional classification is its intuitive interpretation. It is a fair assumption that classifiers are more confident when dealing with familiar test instances whereas their confidence degrade amid novel cases. However, counter-intuitively, learning models' confidence is merely based on the test instance's distance to the decision border and its similarity to training instances has insignificant effect on it. The idea of generalized classification is to remedy this limitation by taking into account the underlying training data distribution when outputing confidence values.

Finally, arguably the final role of learning models should not end with their assigned classification task but rather they should be components of a broader construct. Generalized classification aims to provide models that perform subtasks while not conflicting with other components of the broader construct and also properly responding to unexpected circumstances.

In the canonical classification problem we assume a predefined known set of objects $Y$ that correspond to the class labels available in the training set. In contrast, in generalized classification we assume access only to a subset of the objects in $Y$, $\Theta \subseteq Y$, in the training set while the actual size of $Y$ is unknown. 
Similarly to the definition of risk we define \emph{extended risk} as follows:

Given a non-negative real-valued loss function $L(y',y)$ that measures the loss of predicting $y$ as $y'$ and a function $g:X \rightarrow \{\Theta \cup \varepsilon \}$ that we call a \emph{strong hypothesis}, the extended risk $R^\varepsilon$ of $g$ is defined as: 
\begin{equation} \label{ex risk}
R^\varepsilon(g)=E(L(g(x),y)),
\end{equation}
where $\varepsilon = Y-\Theta$ and thus $D(x,y = \varepsilon) \equiv D(x,y \notin \Theta)$. In other words, a strong hypothesis is a conventional hypothesis when $x$ is an example of a known concept to the model, and a constant mapping $g(x)=\varepsilon$ otherwise. Informally, $\varepsilon$ is an augmented member to the set of known classes that represents all the unknown concepts and a strong hypothesis may assign a test instance to this set by outputting very low probability of inclusion to all known classes. This approach contrasts with a conventional hypothesis that assigns any test instance to one of the known classes without an unknown option. 

\subsection{Generalized Classification and Fooling}
Deactivation of output units for points that are not in the proximity of training data in the initial experiments in the two-dimensional space of the two-circle domain suggested that COOL networks are able to provide strong hypotheses. Further experiments then showed that they can significantly reduce fooling. It is worth highlighting the relationship between strong hypotheses and fooling. In particular, to fool the learning model $g$ one should find an instance $x \sim D(x|y\notin \Theta)$ such that $g(x) \in \Theta$. However, by definition, if $g$ is a strong hypothesis with low extended risk then it tends to assign such an $x$ to $\varepsilon$. Therefore, the closer a learning model is to a strong hypothesis with low extended risk the less likely it is to be fooled.
Note that in this discussion we distinguish \emph{adversarial examples}, which are drawn from $D(x|y \in \Theta)$ but not classified correctly, from arbitrary fooling (false positive) examples. Also note that the same proposition is not valid for a hypothesis that is not strong because:
\begin{itemize}
	\item by definition it has to assign $x$ to a member of $\Theta$ and
	\item there is no obligation imposed on the output of a hypothesis by the loss function when $x \sim D(x|y\notin \Theta)$.
\end{itemize}

\subsection{Separable Concept Learning}
Unfortunately it is not straightforward to formulate the generalized classification problem using an approximation of extended risk through samples in the training set because we have only partial knowledge about $Y$. However, it is possible to evaluate the performance of a strong hypothesis in the framework of classical statistical learning. This evaluation consists of two components: (1) model performance on instance $x$ when $x \sim D(x|y\in \Theta)$, and (2) when $x \sim D(x|y=\varepsilon)$; we call the latter the \emph{inhibition} ability of the model because it must inhibit its tendency to activate at least one output class. While the first component can be evaluated by any classification metric, e.g.\ classification accuracy, for the inhibition ability we introduce the idea of \emph{separable concept learning} (SCL). The advantage of this new metric over previous evaluation procedures applied in open set recognition \citep{openset} and object detection tasks is that it does not rely on any threshold and measures the rejection ability of the model implicitly, which simplifies the evaluation procedure and reduces the processing time significantly.

Intuitively, a strong hypothesis should be able to learn different concepts separately from each other because it is aware of the limits of its own knowledge. As a result, these models can learn new nonoverlapping concepts simply by accepting new output nodes from a separately trained strong hypothesis. In other words, it is possible to train several strong hypotheses over different concepts and merge them together to get a single model over the whole set of concepts. This capability in turn makes \emph{knowledge transfer} possible. Imagine a situation where one has two models to recognize, \{people, animals, plants\} and \{sea, ground, sky\}, respectively. Then a model that finds people in the sea can be simply built by applying a conjunctive operator to the corresponding output nodes in the two models. In a similar manner disjunctive concepts (and concept complements) can be built based on previously learned concepts. Another closely related subject is \emph{incremental concept learning} (ICL), where concepts are presented to the learning model incrementally. Separable concept learning can be conceived as the extreme case of ICL where after learning a few concepts their training instances diminish from the memory. Interestingly, ICL and SCL are more reminiscent of human learning than the current full concept learning approach. Furthermore, in most realistic applications the size of $Y$ is unknown at the beginning, or could grow over time, which makes ICL a practically important subject in supervised learning. 

Formally, to evaluate the inhibition ability of a model, let $concat$ be the concatenation operator and $\rho$ a partition of $\Theta$ such that every subset in $\rho$ is at least of size $2$ (contains at least two concepts from $\Theta$). The \emph{extended empirical risk} of a strong hypothesis $g$ with respect to $\rho$, $R^{\rho}_e(g)$, is then defined as:
\begin{equation} \label{metric}
R^{\rho}_e(g)=\frac{1}{n}\sum_{i=1}^{n}L(\mathrm{concat}_{\forall k \in \rho}(g_k(x_i)),y_i),
\end{equation}
where we assume each strong hypothesis $g_k$ is a learning model trained on the subset of training set samples whose classes are included in the corresponding $\rho$. Note that in case of $\rho=\{Y\}$, equation.~\ref{metric} will reduce to the classification accuracy measure. In addition, merging outputs of different models usually necessitates having outputs in the form of probability values.

Abstractly, while comparing two learning models, $L_1$ is strictly better than $L_2$ if $\forall \rho R^{\rho}_e(L_1)\leq R^{\rho}_e(L_2)$. However, the possible ways of partitioning $\Theta$ grows exponentially with the size of $\Theta$. Thus in practice an approximation is used by sampling from all the possible partitions.

Informally, to apply equation.~\ref{metric}, one can train several components of a SCL model on different subsets of the set of classes and measure the performance of the whole (merged) model over the test set. This approach provides a simple procedure to evaluate different learning models in terms of acting as a strong hypothesis.
\section{Separable Concept Learning Experiment}
To evaluate the feasibility of SCL and consolidate the results from the fooling section, a number of new experiments are conducted with the MNIST and CIFAR-10 datasets. The general scenario is to train multiple networks on different partitions of the training classes and later merge them to evaluate the performance of the whole group, called the \emph{assembly}, on \emph{all} training classes. 

During training a separate validation set (with instances taken only from the same subset of classes as the corresponding training set) is applied to select the best model. As a result, the models are trained completely unaware of other classes and there is no possibility of overfitting. Therefore, both the training set and a separate test set can be applied to measure the performance of the final assembly. 

Because models usually have very high accuracy on their training data, measuring the accuracy on the (combined) training set is a good way to measure the inhibition ability of the models. Also, the test set (of all classes combined) can indicate how well the models can generalize their familiar concepts, as well as inhibit unfamiliar concepts.

On MNIST, the results show that the COOL mechanism can dramatically improve the ability of NNs to learn subsets of concepts separately, both on training and test sets. On the other hand, the CIFAR-10 results with COOL suggest a significant performance improvement only on the training data. We attribute the inconsistent results on test and training data of CIFAR to the significant difference between the distribution of the two sets (i.e.\ training and testing). Overall these results further support that COOL networks are able to successfully find a fit region for each class even in high-dimensional spaces.

\subsection{Experimental Results}
Two CNN architectures are compared that are identical other than their output layers\footnote{The source code for all the experiments in this section can be found at \url{https://github.com/ndkn/fitted-learning}}. Five instances of each variant are then trained on different subsets of MNIST. More specifically, we partition the training set $T$ into five subsets, $T_1$, $T_2$, $T_3$, $T_4$, and $T_5$, where $T_1=\{(x,y)\in T\mid y \in \{0,1\} \}$, $T_2=\{(x,y)\in T\mid y \in \{2,3\} \}$, etc. Then we train a separate CNN from each architecture on $T_i; i\in \{1,2,3,4,5\}$. In other words, $\rho=\{T_1,T_2,T_3,T_4,T_5\}$ in equation~\ref{metric}.     

Table~\ref{mnistscl} shows the average training and test accuracy (equation~\ref{metric}) of the conventional CNN versus COOL CNN over ten different runs of the same experiment. Here the COOL models significantly outperform conventional CNNs with about an $8\%$ improvement on both train and test set accuracies. 

This table also shows the accuracy of MLP architectures on the same task, where an improvement in performance of about $17.8\%$ is achieved simply by applying COOL. Interestingly, the COOL MLP (without convolution) outperforms the conventional CNN on this task.

In the next experiment, in a similar manner, CIFAR-10 is partitioned into five subsets to evaluate the performance of COOL and regular NNs in SCL. CIFAR-10 consists of natural images and is more challenging than MNIST with the best test error (without data augmentation) using simple CNN architectures well above $10\%$. 

Another interesting feature of natural images is they yield a lower median confidence score (output unit activation) on validation/test images \citep{nguyen}, which implies that many test points are within the regions close to the decision borders. Even though COOL is designed to capture the data distribution if provided with sufficient training data, it is interesting to see its performance when the training data is not enough to result in a confident learning model.

Table~\ref{mnistscl} includes the performance of a variety of architectures on CIFAR-10. Here the COOL model is able to significantly improve the training accuracy in SCL though it reduces the test accuracy slightly. 

It is important to note that these experiments apply batch-norm layers after convolutional layers. Without such batch-normalization, the results show an even bigger gap between COOL and regular CNN. Those additional results are included in the appendix.

This result reflects the fact that the model is more confident about the training data and shows that COOL is therefore able to capture the underlying data distribution more accurately than regular NNs. However, because the test data are largely different from training data, the need for generalization is more prominent and thus the regular networks slightly outperform COOL. As expected, if we replace the product with the summation in the inference phase of COOL (aggregation operator in table~\ref{mnistscl}), the COOL acts more like a regular NN and the test accuracy matches that of regular NN.  

\begin{table}[t]
\caption{Generalized classification accuracies for regular and COOL NNs on MNIST and CIFAR-10 data sets. The aggregation operator (the operation applied in the inference phase of COOL) for each COOL architecture is indicated within parentheses.}
\label{mnistscl}
\vskip 0.15in
\begin{center}
\begin{small}
\begin{sc}
\begin{tabular}{lcccr}
\hline
Data set & Architecture & SCL Accuracy (train) \\
\hline
MNIST    &regular MLP& 64.4$\pm$ 0.4 (65.5$\pm$ 0.7) \\
MNIST     & COOL MLP($\times$)& \bf{82.2$\pm$ 0.8} (\bf{83.3$\pm$ 0.8})\\
MNIST      & COOL MLP($+$)& 64.9$\pm$ 0.5 (65.6$\pm$ 0.4)\\
MNIST & regular CNN& 79.4$\pm$ 1.3 (80.5$\pm$ 1.5)\\
MNIST    & COOL CNN($\times$)& \bf{87.6$\pm$ 1.7} (\bf{88.7$\pm$ 1.7})\\
MNIST    & COOL CNN($+$)& 82.1$\pm$ 2.1 (83.2$\pm$ 2.4)\\
CIFAR-10 & regular CNN& 48.6$\pm$ 1.2 (64.2$\pm$ 2.6)\\
CIFAR-10     & COOL CNN($\times$)& 47.1$\pm$ 1.1 (\bf{74.3$\pm$ 3.4})\\
CIFAR-10      & COOL CNN($+$)& 48.7$\pm$ 0.9 (66.4$\pm$ 1.5)\\
\hline
\end{tabular}
\end{sc}
\end{small}
\end{center}
\vskip -0.1in
\end{table}

Overall, these results suggest that COOL can significantly improve the inhibition ability (rejection rate) of learning models, which in turn leads to a more accurate representation of the knowledge embedded in the dataset and robustness of the learned concepts.

\section{One-class Neural Network}
Based on the COOL architecture and following a similar approach to a one-class SVM, this section proposes one-class neural networks, yet another possibility created by COOL. This alternative one-class learning model enables direct application of deep learning methods in the field of one-class recognition. The experiments below suggest that this approach is promising and can allow neural networks to enter the problem domain of estimating the support of probability distributions.
\subsection{Approach}
To capture the underlying distribution of a set of unlabeled data, \citet{oneclasssvm} proposed the one-class SVM, where the origin is treated as the \emph{only} member of the second class. The algorithm then tries to find a maximum margin between the mapped features of the instances of the first class and the origin. To the best of our knowledge, this version and its extensions are the only off-the-shelf methods for capturing a notion of data distribution in high-dimensional spaces (density estimation techniques are usually not effective in high-dimensional spaces because of the curse of dimensionality \citep{scott}) and is applied to anomaly detection, among other tasks.

With a COOL network, one-class neural networks can be constructed in a similar manner by including one (or more) instances that are not within the one single class. In other words, following the same notation introduced in the previous section, the member (or members) of the second class are drawn from $D(x|y \notin \Theta)$ where $|\Theta |=1$. Algorithm \ref{oneclassalg} summarizes the procedure. 
\begin{algorithm}	
	\KwIn{The underlying probability distribution of the non-members of the single class $D'(X)$ = $D(x|y \notin \Theta)$, number of samples from second class  $k$, the set of available instances $T$}
	\KwOut{	COOL neural network $N$}	
	\textbf{step 1:  } form set $\alpha$ by sampling $k$ instances from $D'(X)$\par	
	\textbf{step 2:  } resize $\alpha$ to match the size of $T$ by sampling uniformly from $\alpha$\par
	\textbf{step 3:  } $T \gets T \bigcup \alpha$\par	
	\textbf{step 4:  } train $N$ on $T$\par
	\caption{The one-class neural network algorithm. }
	\label{oneclassalg}
\end{algorithm}
\subsection{Experiments}    
The experiments that follow demonstrate the abilities of one-class neural networks. In the first experiment a similar problem to the two-circle problem in visualization experiments section is introduced. More specifically, to form the training set, points are sampled uniformly within a circle $C$. To form an instance from the second class, an arbitrary point not within $C$ is selected. Figure~\ref{oneclass} depicts the activation map of the two output aggregates (DOO $=5$) assigned to class one and two, respectively. For comparison, the result of the same experiment is also shown when typical neural networks are applied instead of COOL ones. These plots support the validity of the proposed one-class neural network in low-dimensional spaces where the regular neural network fails.

\begin{figure}[ht]
	
	\begin{minipage}{0.50\columnwidth}
		\centerline{\includegraphics[width=\columnwidth]{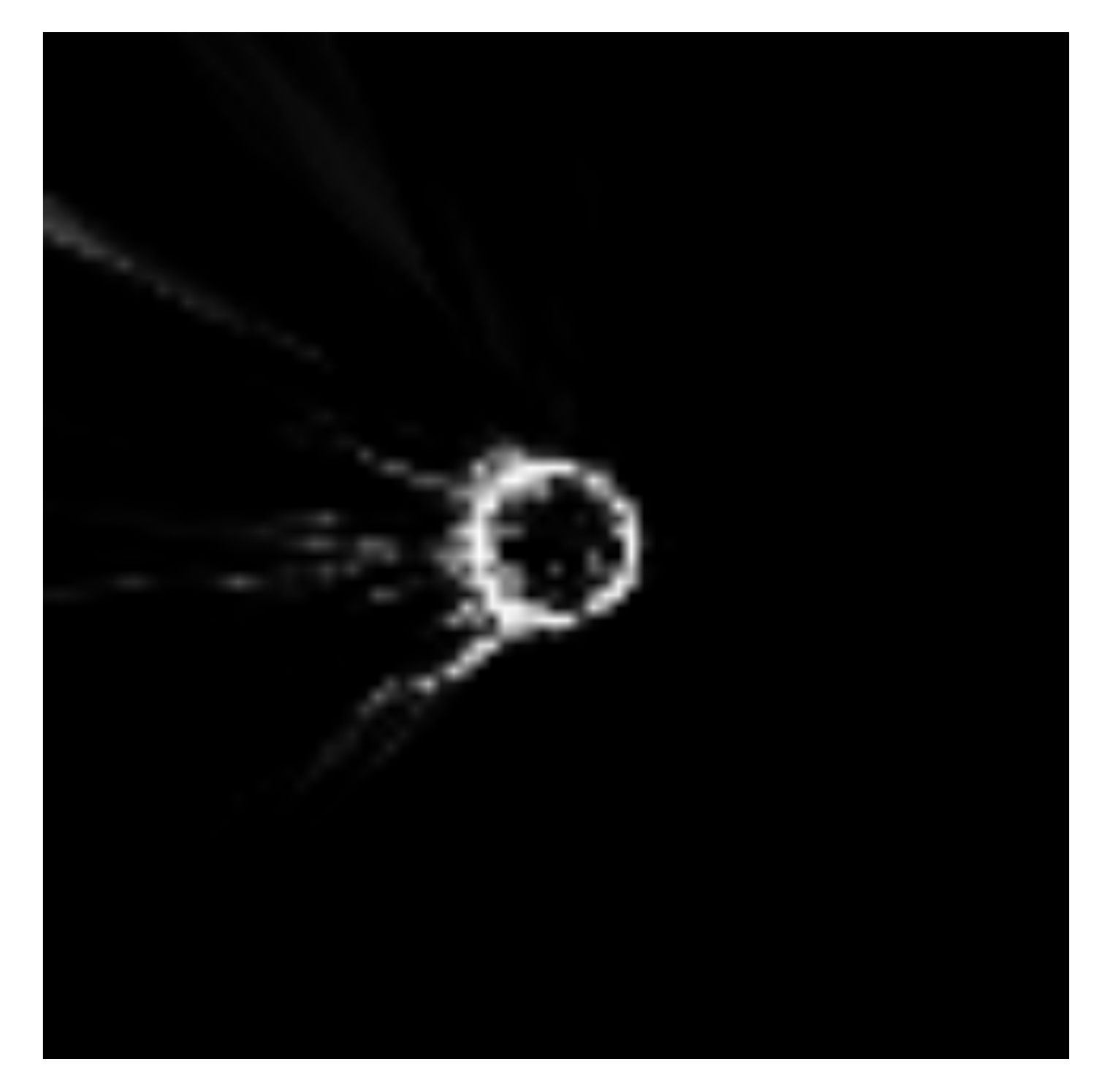}}
		\captionsetup{labelformat=empty,skip=8pt,justification=raggedright,labelsep=newline,singlelinecheck=false}
		\caption{(a) COOL class $1$ output\\}
	\end{minipage}%
	\begin{minipage}{0.49\columnwidth}
		\centerline{\includegraphics[width=\columnwidth]{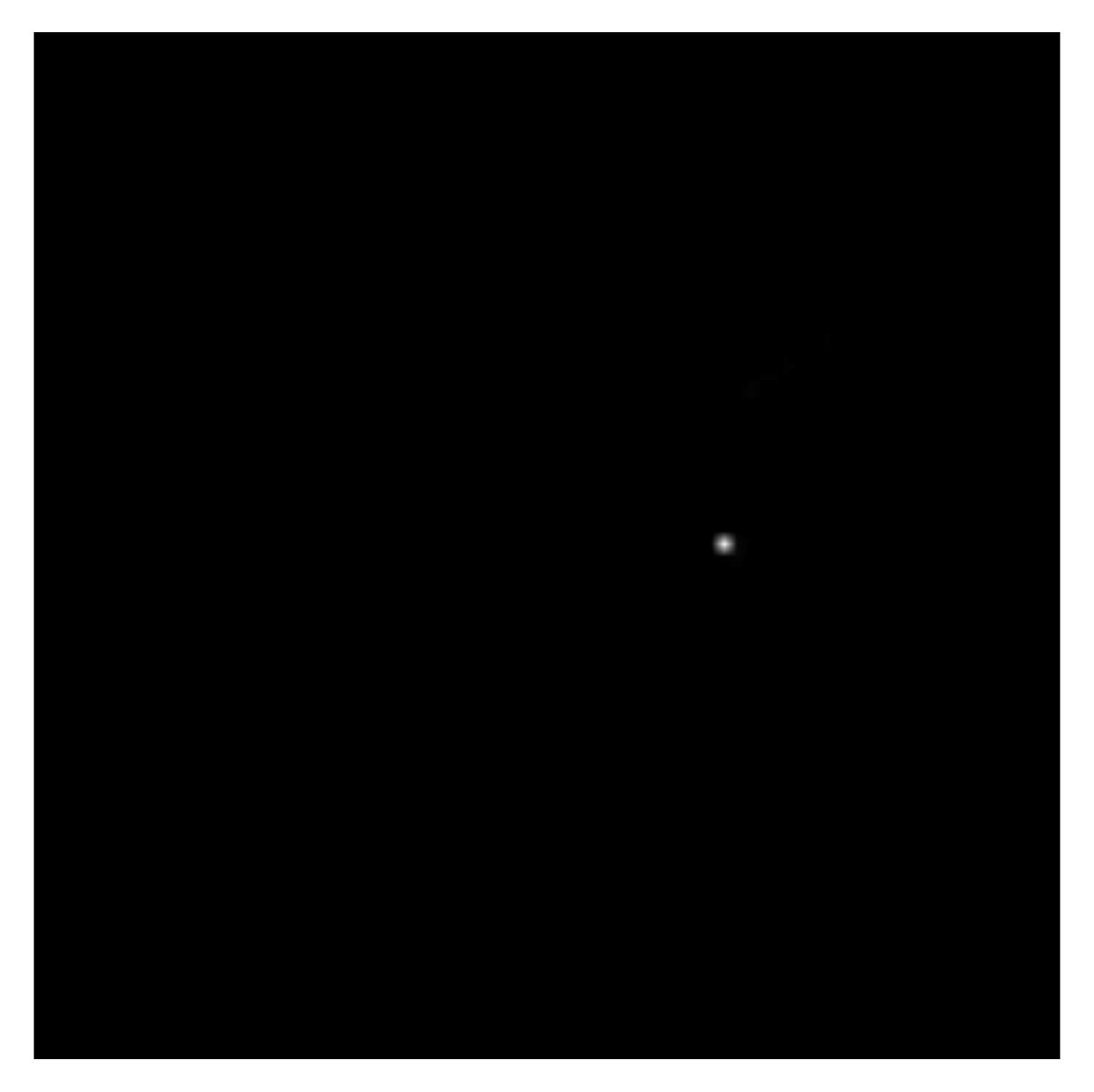}}
		\captionsetup{labelformat=empty,skip=8pt,justification=raggedright,labelsep=newline,singlelinecheck=false}
		\caption{(b) COOL class $2$ output\\}
	\end{minipage}	
	\begin{minipage}{0.50\columnwidth}
		\centerline{\includegraphics[width=\columnwidth]{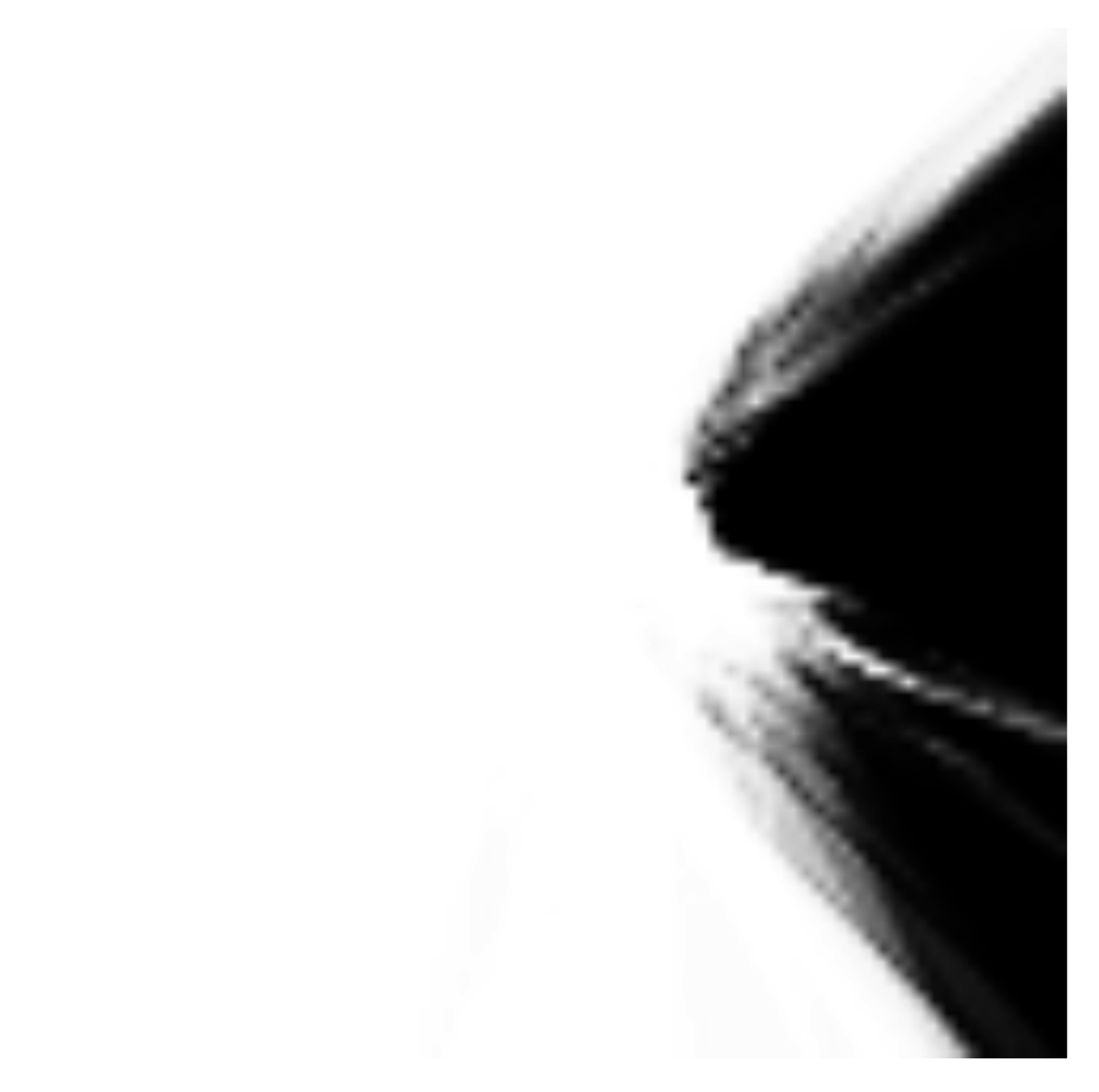}}
		\captionsetup{labelformat=empty,skip=8pt,justification=raggedright,labelsep=newline,singlelinecheck=false}
		\caption{(c) Regular neural network\\ class $1$ output}
	\end{minipage}%
	\begin{minipage}{0.49\columnwidth}
		\centerline{\includegraphics[width=\columnwidth]{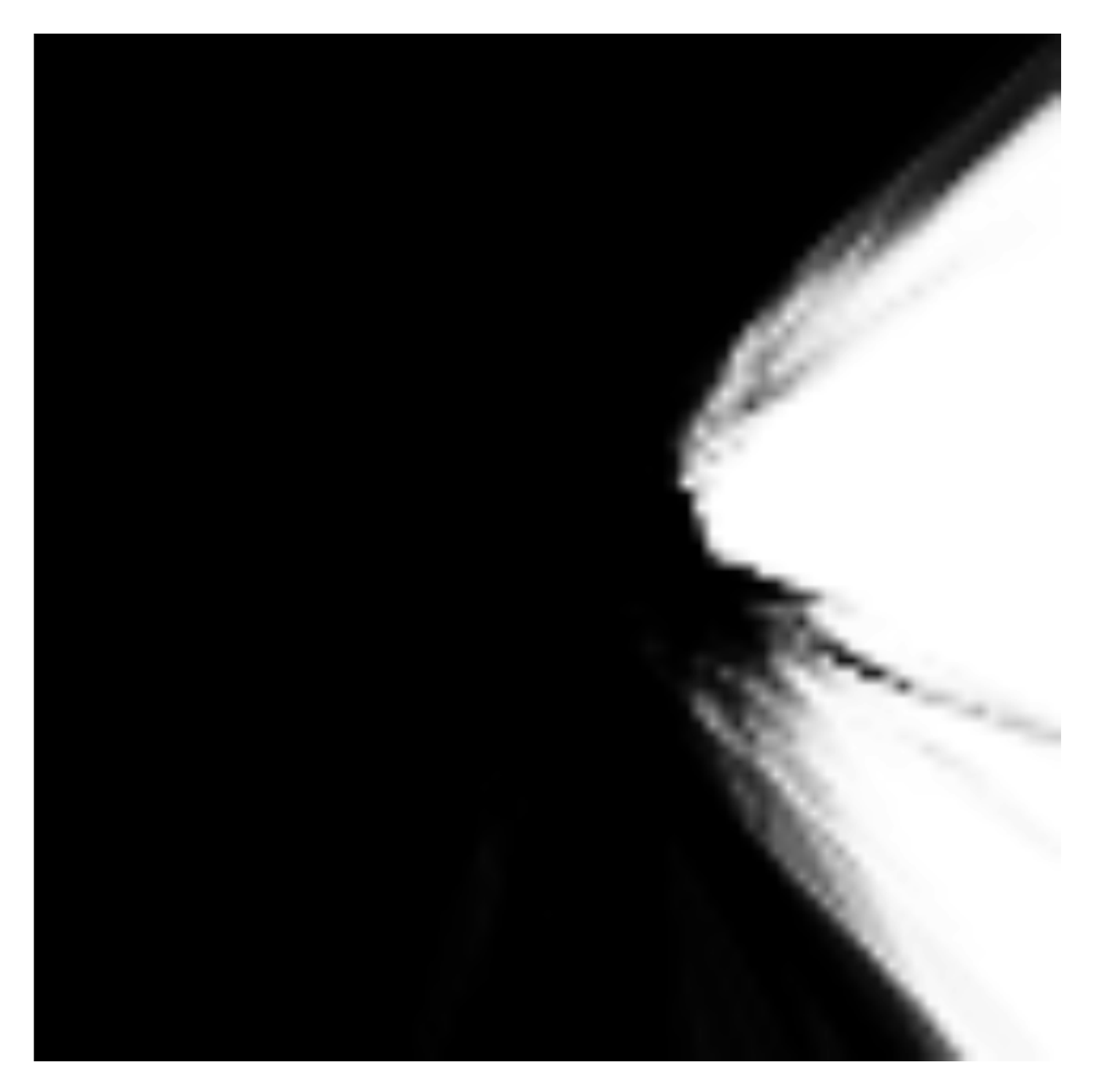}}
		\captionsetup{labelformat=empty,skip=8pt,justification=raggedright,labelsep=newline,singlelinecheck=false}
		\caption{(d) Regular neural network\\ class $2$ output}
	\end{minipage}
	\setcounter{figure}{11}	
	\caption{\textbf{One-class neural network visualizations.} Visualizations (a) and (b) depict the activation maps of the output aggregates on a one-class version of the visualization problem. These plots show how accurately one-class neural networks can capture the distribution of data in the only present class (a). For comparison, (c) and (d) depict the same activation maps when the COOL component in the one-class neural network is replaced with conventional neural network outputs. }
	\label{oneclass}
\end{figure}

Next, MNIST tests the performance of one-class neural networks in a high-dimensional problem. In this experiment ten separate one-class neural networks are trained to learn the ten different classes of MNIST. Furthermore, one image with random pixel values is supplied as the only instance of the second class for all of these ten models. That way, each model is supposed to capture the data distribution of a single digit. Later, to test instance $x$, it is fed to all ten models and then assigned to the model with the maximum probability of inclusion (i.e.\ highest activation). In this manner, each model is treated as an unnormalized probability distribution function.

\begin{figure}[!t]
	\vskip 0.1in
	\begin{center}
		\centerline{\includegraphics[width=\columnwidth]{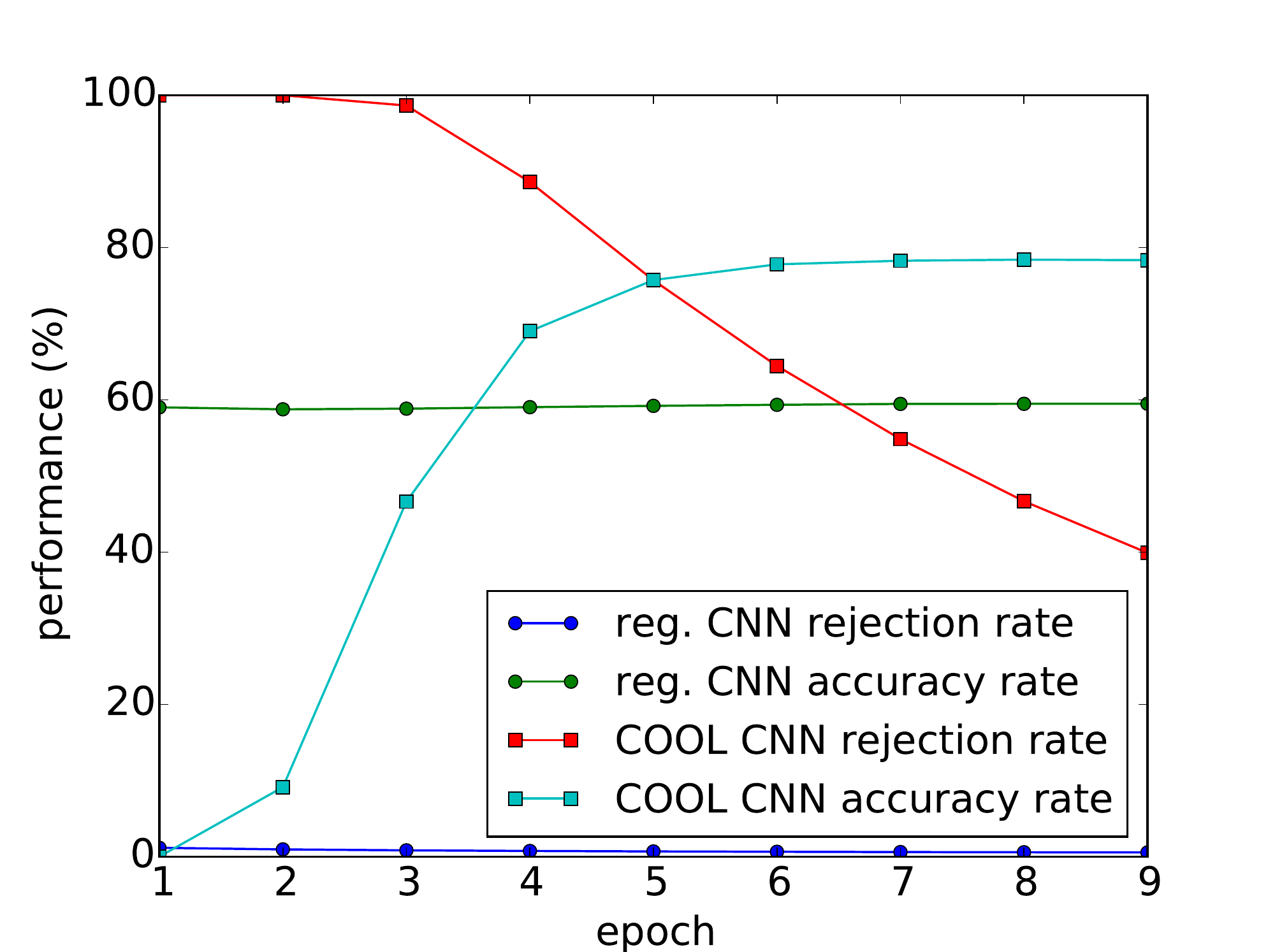}}
		\caption{\textbf{Average accuracy vs. rejection rate in one-class experiment.} These plots show the average accuracy and rejection rates of the ten models over the test instances (from all $10$ classes) in the first ten epochs of the MNIST one-class experiment when $0.5$ is applied as the threshold value. The plots show that though regular CNNs reach some level of accuracy, their correct rejection rate is too low to make them an interesting one-class model.}
		\label{accvsrej}
	\end{center}
	\vskip -0.1in
\end{figure}

The main results are average test accuracies of $75.7\%$ and $60\%$ over five runs achieved by COOL (DOO $= 80$) and the conventional CNN, respectively. However, because classification accuracy is solely based on the maximum activation (probability of inclusion), it is not a good representative of the magnitude  of probability values. Furthermore, in one-class learning problem the performance of the model is based on both its correct detection \emph{and} rejection, which can be easily violated by classic discriminative model (due to the high data imbalance). Thus, to better demonstrate the difference between COOL and conventional CNNs in this experiment, figure~\ref{accvsrej} depicts the average accuracy and rejection rate of COOL versus regular CNNs over different epochs when $0.5$ activation is set as the threshold for rejection (and minimum required activation in calculating accuracy). In other words, if a model's activation for test instance $x$ is below $0.5$, it is interpreted as rejecting the inclusion of $x$ to the corresponding class (the class associated with the learning model). Notice in figure~\ref{accvsrej} that the rejection rate of the regular CNN is flat near the implausible value of zero, while that of COOL drops to about $60\%$ when its accuracy rate flattens. Interestingly, \citet{Jain2014} report in their supplemental material an accuracy of about $52\%$ for one-class RBF SVM in a similar experiment conducted on $6$ classes of MNIST.\footnote{\citet{Jain2014} did not include the rejection rate in their report.} 
These results further affirm the advantage of COOL networks and their ability to capture the underlying distribution of data. Moreover, they support the effectiveness of the proposed one-class neural networks. 
\section{CIFAR-10 and CIFAR-100 Experiments}
So far we have focused on the ability of COOL to capture the underlying distribution of data and generalize adequately. Another potential advantage of COOL is its ability to train larger architectures. As noted earlier, COOL in effect trains an exponential number of competing models together. An intriguing consequence is that it is possible to train very large architectures without overfitting. As results in this section will show, this property alone can sometimes lead to performance improvements. This is usually the case when the networks are very wide.

Another important question is the rate of convergence in COOL. Is the aggressive competition in the output layer of these networks a hindrance to convergence, or, conversely, does it lead to more robust updates of parameters in the training phase and therefore accelerated learning?

In this section we conduct experiments with different network architectures on the CIFAR-10 and CIFAR-100 datasets to evaluate the effect of COOL on the rate of convergence and on performance. Using Torch\footnote{\url{http://torch.ch/}}, the architectures in this section apply the latest techniques in deep learning to check their synergy with COOL, though again the learning procedure is kept as simple as possible.

\subsection{CIFAR-10 Experiment}
In this experiment a CNN with five convolutional layers followed by two fully-connected layers is trained on the first $45$,$000$ training instances of the CIFAR-10 dataset. Each hidden layer is followed by a batch normalization layer and ReLU is applied as activation functions in all layers. More details are in the Appendix.

Figure~\ref{cifar10} depicts the average classification accuracy over ten runs of the COOL network (with DOO $= 20$) versus ten runs of a conventional network with the same architecture, except for the last layer. The main result is that COOL networks converge more than two times faster while performing significantly better ($t$-test produces a $p$-value$<0.001$). Preliminary experiments with a variety of different architectures yielded similar results.

\begin{figure}[ht]
	\vskip 0.1in
	\begin{center}
		\centerline{\includegraphics[width=\columnwidth]{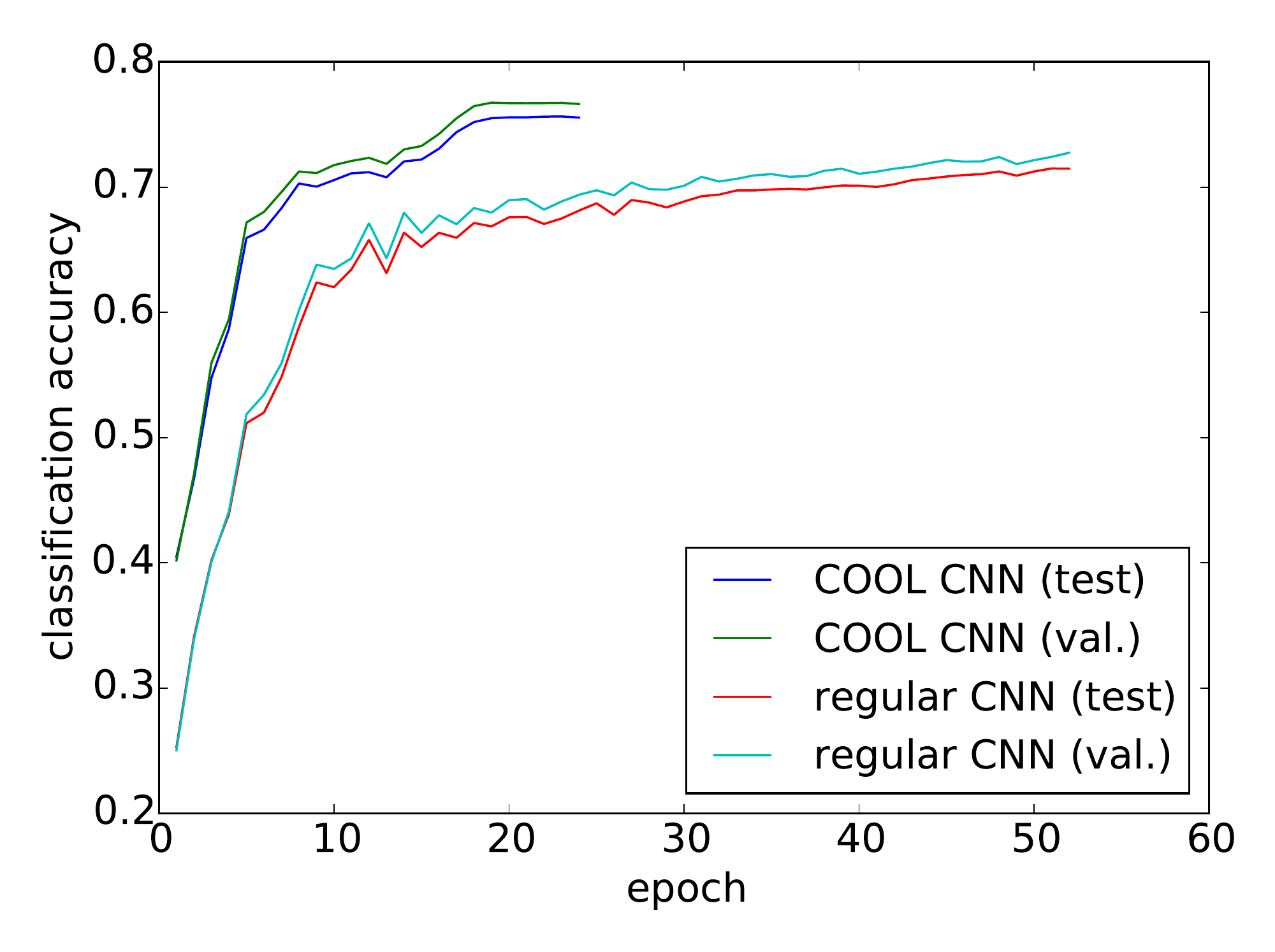}}
		\captionsetup{skip=5pt}
		\caption{\textbf{Average classification accuracy of COOL and a conventional CNN with the same architecture on CIFAR-10 over $10$ runs.} COOL speeds up learning significantly while also improving test accuracy. This result suggest more robust gradient updates during training.}
		\label{cifar10}
	\end{center}
	\vskip -0.1in
\end{figure}

\subsection{CIFAR-100 Experiment}
In this experiment a more comprehensive architecture is applied to the CIFAR-100 dataset. The unique characteristics of this experiment are the application of dropout and a light augmentation of the dataset according to the same procedure as in \citet{elu}, where each image is padded by four zero pixels at all borders, randomly cropped to extract a $32\times32$ image, and also randomly horizontally flipped. To keep the training procedure simple, the dropout rate is kept constant and no momentum or weight decay is applied. More details are in the Appendix.

\begin{figure}[ht]
	\vskip 0.1in
	\begin{center}
		\centerline{\includegraphics[width=\columnwidth]{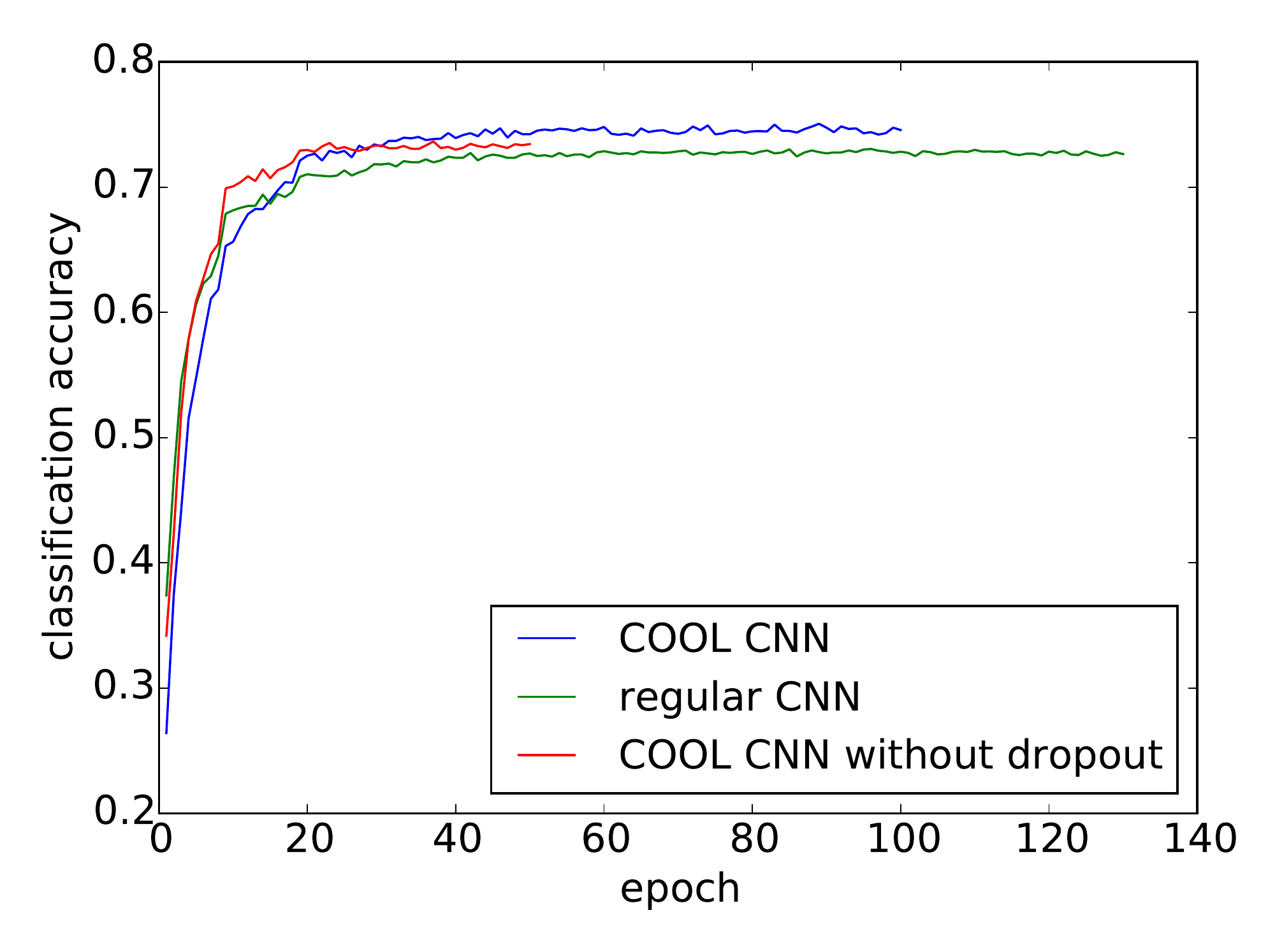}}
		\captionsetup{skip=5pt}
		\caption{\textbf{Performance of COOL and non-COOL CNNs with the same architecture on CIFAR-100.} These plots show the synergy of COOL with dropout.}
		\label{cifar100}
	\end{center}
	\vskip -0.1in
\end{figure}

The classification accuracy of the conventional versus the COOL neural network without dropout over different epochs of the run are shown in figure~\ref{cifar100}. These results suggest, as expected, that dropout slows down the learning process significantly while improving performance. However, they also show once again that COOL converges faster and to a higher level of accuracy. For comparison, Table~\ref{cifar100results} lists some of the state-of-the-art results on CIFAR-100, including AlexNet \citep{alexnet}, DSN \citep{dsn}, NiN \citep{NiN}, Maxout \citep{maxout}, All-CNN \citep{all-cnn}, Highway Network \citep{highway}, Fractional Max-Pooling \citep{fractional}, and ELU-Network \citep{elu}. 

\begin{table}[t]
	\caption{\textbf{Comparison of COOL network and some recent successful CNN architectures on CIFAR-100.}}
	
	\label{cifar100results}
	\centering
	\begin{tabular}{ll}
		\toprule
		\cmidrule{1-2}
		Network     & CIFAR-100 (test accuracy \%)  \\
		\midrule
		AlexNet & 54.2   \\
		DSN     & 65.43    \\
		NiN     & 64.32        \\
		Maxout     & 61.43        \\
		All-CNN     & 66.29       \\
		Highway Network     & 67.76        \\
		Fract. Max-Pooling     & 72.38        \\
		ELU-Network     & \textbf{75.72}       \\
		\textbf{COOL CNN w/o dropout}     & 73.5     \\
		\textbf{COOL CNN with dropout}     & 74.72  \\
		CNN with dropout (with \\same arch. as COOL CNN)  & 72.79  \\
		\bottomrule
	\end{tabular}
\end{table}
\section{Discussion and Future Work}

Formulating the notion of generalized classification helps to illuminate the rigidness of more conventional classification, which plays a central role in many of the most impactful benchmark results in deep learning. 
Generalized classification is arguably more closely aligned with a natural setting in which the set of classes or concepts is fluid and subject to expansion.  It also enshrines the notion of the unknown as a fundamental ingredient in the framework of learning, an intuitive idea for humans where deep networks nevertheless can falter (as in fooling).  Fitted learning thereby becomes a natural response to the problem of generalized classification, lighting a path towards new approaches and architectures.

One such new approach is the COOL network introduced in this paper.  While the introductory experiments disclosed here are only an initial hint and more investigation will be necessary, by taking a significant step towards better fitted learning, COOL shows that the mechanisms that can enable a more dynamic form of learning are not necessarily complex or intricate.  Rather, overrepresentation in the output layer combined with competition precipitates a partitioning of the input space among multiple overlapping models simultaneously, yielding a better fit to the data in aggregate. In this way, COOL in effect produces an ensemble of an exponential number of learning models, which is reminiscent of dropout. Even outside of separable learning and one-class neural networks, from the results in CIFAR-10 and CIFAR-100, the advantages of this modification seem to extend to conventional classification.  It is an interesting question whether the main insights in COOL might bolster the case against single-neuron concepts in general, including in the brain.  

At a practical level, the ability to add new classes to a model without the need to retrain opens the door to a new class of continual lifetime learners that expand their repertoire of concepts indefinitely.  In this way, all the power of deep learning can be naturally brought to bear on lifetime learning, an exciting proposition.  While a simple form of such continual learning is demonstrated in this paper where different modules learn different classes independently before later being combined, more sophisticated variations on this idea are conceivable and likely to emerge in the near future.  For example, as new modules that address new concepts are added they can also potentially share hidden layers with old modules, allowing a safe form of expansion that also builds on accumulated knowledge.  Tuning such hidden knowledge as the aggregate learner expands also offers a breadth of possible investigations.  Furthermore, the potential benefits to recurrent neural networks remain unexplored yet promising.

In short, the fitted learning of COOL creates new possibilities and mitigates awkward problems (such as fooling) because it takes a qualitative step towards a more natural kind of learning.  That is, COOL is not just about more accurate classification, but about a different \emph{perspective} on classification, towards generalized classification.  While progress sometimes follows a path of quantitatively increasing performance, such as recent advances in accuracy in deep learning, often it also requires a new perspective to reimagine the status quo in a new light.   While the results from COOL in this paper are too preliminary to declare victory in the problem of generalized classification, our hope is that COOL will provide such a productive shift in perspective that will at least set us down that path, and one day yield learners more in the dynamic and continual spirit of humans.
\section{ Acknowledgments}
Special thanks to Omid Kardan for helpful discussions and comments.
\medskip

\section{Appendix}
This appendix gives a more detailed description of the experiments in the paper. The COOL architecture introduces two new hyperparameters, namely the DOO and softness. The softness is always $1.0$ except in the single visualization experiment that is designed to show its effect on the decision border. The DOO is typically $5$, $10$, or $20$ depending on the type of dataset and problem because it is basically the number of input space partitions. For example, the two-circle problem with only two-dimensions does not in effect ask for as many partitions by COOL member units as in MNIST with $28\times28$ inputs. For tasks such as the one-class neural network a higher value makes sense because of the high data imbalance (and therefore tendency to overgeneralize) but in general we do not optimize this parameter systematically. Most experiments are aimed at simple architectures, except for CIFAR-10 and CIFAR-100, where state-of-the-art architectures are tested. 
\subsection{Visualization Experiments}

The two-circle problem consists of $1$,$258$ points uniformly sampled from two concentric circles with different radii. Table \ref{architectures} lists the different feed-forward architectures applied throughout this paper to generate the visualizations\footnote{The source code for visualizations in Figure \ref{logarithmicspiral}, including the architectures, are provided at \url {https://github.com/ndkn/fitted-learning}. }. 
\begin{table}[!htbp]
	\caption{\textbf{The architectures applied in the visualization experiments.}}
	
	\label{architectures}
	\centering
	\begin{tabular}{ll}
		\toprule
		\cmidrule{1-2}
		Experiment & Network Architecture     \\
		\midrule
		figure $3$a & 2,200,2   \\
		figure $3$b     & 2,400,300,200,2    \\
		figure $4$a-b      & 2,400,300,10        \\
		figure $4$c-d     & 2.400,300,2        \\
		figure $5$a     & 2,400,300,20       \\
		figure $5$b   & 2,400,300,4        \\
		figure $6$a     & 2,200,10        \\
		figure $6$b     & 2,400,350,300,250,200,10  \\
		figure $8$     & 2,800,300,100,15     \\
		figure $9$     & 2,600,200,10  \\
		figure $12$a-b   & 2,600,200,10  \\
		figure $12$c-d   & 2,600,200,2  \\
		\bottomrule
	\end{tabular}
\end{table}
\subsection{Experiments with MNIST}
The MNIST dataset is a popular machine learning benchmark consisting of $60$,$000$ training and $10$,$000$ $28\times28$ digit test images. In all the experiments the first $50$,$000$ training images are used as the training set and the rest as validation. The neural network architectures are kept simple (most CNN architectures are inspired by LeNet-5 \citep{mnist}) and all the hidden units apply a saturating activation function (logistic). In the preprocessing step we simply transform all pixel intensity values to the $[0,1]$ interval by dividing them by $255$. Finally, the learning rate is constant and the weight initialization follows \citet{xavier}.
\subsubsection{Fooling Experiment}
In this experiment we first train two CNNs with a $[Conv(1\rightarrow20,5\times5), maxPooling(2\times2), Conv(20\rightarrow50,5\times5), maxPooling(2\times2), fullyConnected(50\rightarrow400), fullyConnected(400\rightarrow10\times DOO)]$ architecture, where the DOO is $1$ for the regular and $10$ for the COOL network, respectively. The learning rate is $0.001$ and the training continues for at most $200$ epochs.

Next are the trials on both trained networks. Each trial consists of (1) generating a $28\times28$ image with pixels sampled from the uniform distribution in the $[0,1]$ interval, (2) training a FGN (learning rate $= 0.00001$) consisting of a single fully-connected layer that transforms this image into a $28\times28$ fooling (false positive) image with all pixels in the $[0,1]$ interval. In the case of the COOL network, the FGN is trained based on the errors backpropagating from member units of a specific class and not the final result after multiplication. As a side note we also tried FGNs with several layers but the single-layer FGN works the fastest while serving the role of constraining the pixel intensities to the valid range of $[0,1]$.
\subsubsection{Separable Concept Learning Experiment}
This experiment evaluates the performance of COOL versus a regular neural network on both CNN and MLP architectures on a different architecture that discussed in the main body of this paper. The main difference is that the CNN model in this experiment does not apply batch-normalization.

Again, in the training phase, five models are trained, each on different pairs of classes. During testing, a test instance is fed to all the five models and the class with the highest activation is selected as the final decision. The validation set includes examples from all the $10$ classes and is the only means to tune the amount of generalization in the final models.

The MLP architecture consists of $3$ hidden layers each with $500$ units and the DOO is one and $10$ for the regular and COOL network, respectively. The training applies a learning rate of $0.005$ and is continued for the maximum of $120$ and $250$ epochs for COOL and regular network, respectively, and the reported MLP results are the average of $2$ runs.

The CNN architecture is $[Conv(1\rightarrow20, 5\times5), maxPooling(2\times2), Conv(20\rightarrow50(10),  5\times5), maxPooling(2\times2), fullyConnected(50\rightarrow400), fullyConnected(400\rightarrow10\times DOO)]$, where to speed up training we apply strides of $2$ to both convolutional layers and the second convolutional layer is mapped randomly to half of the feature maps from the previous layer. Again, the DOO is set to $1$ and $10$ for conventional and COOL network, respectively, and the learning rate is $0.005$. The CNN results are the average of $10$ runs of this experiment.  

\subsection{Results}
\begin{figure}[!t]
	\vskip 0.1in
	\begin{center}
		\centerline{\includegraphics[width=\columnwidth]{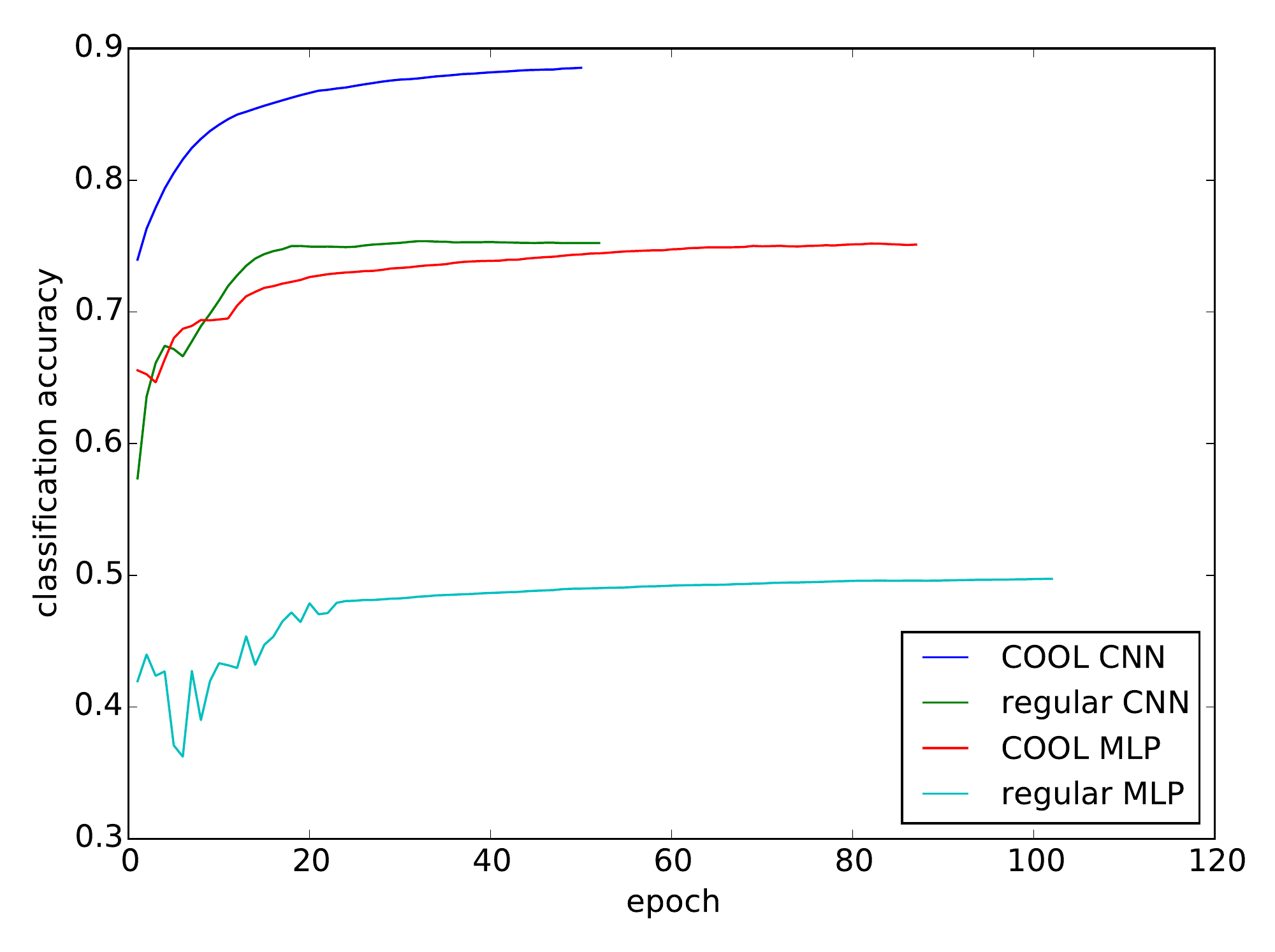}}
		\captionsetup{skip=5pt}
		\caption{\textbf{Test performance of different learning models in separable concept learning in MNIST.} In this experiment the performance of five neural networks, each trained on a different pair of digits from MNIST, is evaluated when they are merged. The main result is the improvement of COOL over conventional neural networks.}
		\label{dcl}
	\end{center}
	\vskip -0.1in
\end{figure}

Figure~\ref{dcl} shows the average test/validation error (equation~\ref{metric}) of the conventional CNN versus ~COOL CNN over epochs. Here the COOL models significantly outperform conventional CNNs with about a $13\%$ improvement in recognition rate (over ten runs). Figure~\ref{dcl} also shows the accuracy of MLP architectures (over two runs) on the same task, where an improvement in performance of about $25\%$ is achieved simply by applying the COOL. Interestingly, the COOL MLP (without convolution) achieves competitive performance with the conventional CNN on this task.

\subsubsection{One-class Neural Network Experiment}
In these experiments the same CNN architecture as in the separable concept learning experiment is applied to train $10$ different models. The DOO for the regular and COOL networks are one and $80$ respectively. We also tried DOOs of $20$ and $50$ but found that higher DOO leads to better results in the trade-off between classification accuracy and rejection rate. 
\subsection{Experiments on CIFAR-10}
The CIFAR-10 dataset consists of $50$,$000$ of training and $10$,$000$ testing $3\times32\times32$ color images from $10$ categories of some animal and transportation media. The validation set is the last $5$,$000$ images of the training images, which are excluded from the training set. For preprocessing, we simply subtract the mean and divide the standard deviation of each pixel location from pixel values. The architecture is:\\
$[Conv(3 \rightarrow 16, 3\times3),Conv(16 \rightarrow 32, 5\times5),maxPooling(2\times2),Conv(32 \rightarrow 64, 3\times3),Conv(64 \rightarrow 128, 5\times5),maxPooling(2\times2),Conv(128 \rightarrow 256, 3\times3),fullyConnected(256 \rightarrow 800),fullyConnected(800 \rightarrow 10\times DOO)]$, \\
where all layers (except for the output and pooling layers) follow batch normalization and apply ReLU as the activation function. The DOO is $1$ and $10$  for regular and COOL networks, respectively and the learning rate is $0.01$ with the batch sizes of $500$.
\subsection{Experiments on CIFAR-100}
The CIFAR-100 dataset is another dataset of small images similar to CIFAR-10 but it contains $100$ categories with the same total number of images. The experiment on CIFAR-100 applies the same preprocessing as with CIFAR-10 but on image derivatives (with respect to $x$) instead of the raw pixel values. The architecture is:\\
$[Conv(3\rightarrow128, 1\times1),Conv(128\rightarrow512, 3\times3), maxPooling(2\times2), Conv(512\rightarrow512, 1\times1), Conv(512 \rightarrow 1024, 3\times3), Dropout(0.1), maxPooling(2\times2), Conv(1024 \rightarrow 1024, 1\times1),Conv(1024 \rightarrow 3072, 3\times3), Dropout(0.2), maxPooling(2\times2), Conv(3072 \rightarrow 3072, 1\times1),Conv(3072 \rightarrow 8192, 3\times3), Dropout(0.3), maxPooling(2\times2), Conv(8192 \rightarrow 8192, 1\times1), Conv(8192 \rightarrow 100\times DOO, 3\times3), maxPooling(2\times2)]$, 
\\where each dropout layer and the last convolutional layer follow batch normalization and ReLU is the activation function. The DOO is $1$ and $20$  for regular and COOL networks, respectively and the learning rate is $0.001$, which is halved every $10$ epochs. 
\end{document}